\documentclass[11pt,twoside]{article}

\usepackage[utf8]{inputenc}
\usepackage{fullpage}

\usepackage{epsf}
\usepackage{fancyhdr}
\usepackage{graphics}
\usepackage{graphicx}
\usepackage{psfrag}

\usepackage{color}

\usepackage{amsthm}
\usepackage{amsfonts}
\usepackage{amsmath}
\usepackage{amssymb}

\newlength{\widebarargwidth}
\newlength{\widebarargheight}
\newlength{\widebarargdepth}

\makeatletter
\long\def\@makecaption#1#2{
        \vskip 0.8ex
        \setbox\@tempboxa\hbox{\small {\bf #1:} #2}
        \parindent 1.5em  
        \dimen0=\hsize
        \advance\dimen0 by -3em
        \ifdim \wd\@tempboxa >\dimen0
                \hbox to \hsize{
                        \parindent 0em
                        \hfil 
                        \parbox{\dimen0}{\def\baselinestretch{0.96}\small
                                {\bf #1.} #2
                                } 
                        \hfil}
        \else \hbox to \hsize{\hfil \box\@tempboxa \hfil}
        \fi
        }
\makeatother

\usepackage{amsmath,amsfonts,amssymb,bm}
\usepackage{framed}
\usepackage{algorithm,algpseudocode}
\usepackage{caption,subcaption,graphicx,rotating,multirow}
\usepackage{hyperref,url}
\usepackage{color,xcolor}
\usepackage{enumitem}
\usepackage{epstopdf}
\usepackage[numbers]{natbib}
\newtheorem{theorem}{Theorem}[section]

\newtheorem{lemma}{Lemma}[section]

\def\BC{\begin{center}}
\def\EC{\end{center}}
\def\BIT{\begin{itemize}}
\def\EIT{\end{itemize}}
\def\BET{\begin{enumerate}}
\def\EET{\end{enumerate}}
\def\BEQ{\begin{equation}}
\def\EEQ{\end{equation}}

\newcommand{\R}{\mathbb{R}}

\newcommand{\trace}{\mathsf{Tr}}

\newcommand{\lbr}{\langle}
\newcommand{\rbr}{\rangle}
\renewcommand{\P}{\mathbb{P}}

\newcommand{\abs}[1]{\left|#1\right|}
\renewcommand{\P}{\operatorname{\mathbb{P}}}
\newcommand{\E}{\operatorname{\mathbb{E}}}

\newcommand{\ObsSet}{\ensuremath{O}}
\newcommand{\numobs}{\ensuremath{N}}
\newcommand{\numlabeled}{\ensuremath{n}}
\newcommand{\graph}{\ensuremath{G}}
\newcommand{\vertex}{\ensuremath{V}}
\newcommand{\edge}{\ensuremath{E}}
\newcommand{\fstar}{\ensuremath{f^*}}
\newcommand{\defn}{\ensuremath{: \, = }}
\newcommand{\usedim}{\ensuremath{d}}
\newcommand{\widgraph}[2]{\includegraphics[keepaspectratio,width=#1]{#2}}

\newcommand{\DIV}{\ensuremath{\operatorname{div}}}
\newcommand{\Gfun}{\ensuremath{\varphi}}
\newcommand{\bandwidth}{\ensuremath{h}}
\newcommand{\Hil}{\ensuremath{\mathcal{H}}}
\newcommand{\Lfun}{\ensuremath{\mathcal{L}}}

\newcommand{\Ker}{\ensuremath{\mathcal{K}}}
\newcommand{\kind}{\ensuremath{k}}
\newcommand{\jind}{\ensuremath{j}}

\newcommand{\vol}{\ensuremath{\operatorname{vol}}}

\newcommand{\fhat}{\ensuremath{\widehat{f}}}

\newcommand{\elltwo}[1]{\ensuremath{\| #1 \|_2}}

\newcommand{\order}{\ensuremath{\mathcal{O}}}

\newcommand{\Ball}{\ensuremath{\mathbb{B}}}

\newcommand{\Fclass}{\ensuremath{\mathcal{F}}}
\newcommand{\Gcomp}{\ensuremath{\mathbb{G}}}
\newcommand{\real}{\ensuremath{\mathbb{R}}}

\newcommand{\pdens}{\ensuremath{\mu}}
\newcommand{\pmax}{\ensuremath{\pdens_{\tiny{\operatorname{max}}}}}
\newcommand{\pmaxsq}{\ensuremath{\pdens^2_{\tiny{\operatorname{max}}}}}

\newcommand{\fbar}{\ensuremath{\bar{f}}} 

\newcommand{\kereig}{\ensuremath{\gamma}}

\begin{document}

\begin{center}

{\bf{\LARGE{Asymptotic behavior of
  $\ell_p$-based Laplacian regularization in semi-supervised
  learning}}}

\vspace*{.2in}

{\large{
\begin{tabular}{ccc}
Ahmed {El Alaoui}$^{\star}$ & Xiang Cheng$^{\star}$ & Aaditya Ramdas$^{\star,\dagger}$ 
\end{tabular}
\begin{tabular}{cc}
 Martin J. Wainwright$^{\star,\dagger}$  &Michael I. Jordan$^{\star,\dagger}$ \\
\end{tabular}
}}

\vspace*{.2in}

\begin{tabular}{c}
Department of Electrical Engineering and Computer Sciences$^\star$, and\\
Department of Statistics$^\dagger$, \\
UC Berkeley,  Berkeley, CA  94720.
\end{tabular}

\vspace*{.2in}


\end{center}

\vspace*{.2in}

\begin{abstract}
Given a weighted graph with $\numobs$ vertices, consider a real-valued
regression problem in a semi-supervised setting, where one observes
$\numlabeled$ labeled vertices, and the task is to label the remaining
ones. We present a theoretical study of $\ell_p$-based Laplacian
regularization under a $d$-dimensional geometric random graph model.
We provide a variational characterization of the performance of this
regularized learner as $\numobs$ grows to infinity while $\numlabeled$
stays constant; the associated optimality conditions lead to a partial
differential equation that must be satisfied by the associated
function estimate $\fhat$.  From this formulation we derive several
predictions on the limiting behavior the $d$-dimensional function
$\fhat$, including (a) a phase transition in its smoothness at the
threshold $p = d + 1$; and (b) a tradeoff between smoothness and
sensitivity to the underlying unlabeled data distribution $P$.  Thus,
over the range $p \leq d$, the function estimate $\fhat$ is degenerate
and ``spiky,'' whereas for $p\geq d+1$, the function estimate $\fhat$
is smooth.  We show that the effect of the underlying density vanishes
monotonically with $p$, such that in the limit $p = \infty$,
corresponding to the so-called Absolutely Minimal Lipschitz Extension,
the estimate $\fhat$ is independent of the distribution $P$. Under the
assumption of semi-supervised smoothness, ignoring $P$ can lead to
poor statistical performance; in particular, we construct a specific
example for $d=1$ to demonstrate that $p=2$ has lower risk than
$p=\infty$ due to the former penalty adapting to $P$ and the latter
ignoring it.  We also provide simulations that verify the accuracy of
our predictions for finite sample sizes.  Together, these properties
show that $p = d+1$ is an optimal choice, yielding a function estimate
$\fhat$ that is both smooth and non-degenerate, while remaining
maximally sensitive to $P$.
\end{abstract}








\section{Introduction}


Semi-supervised learning is a research field of growing interest in
machine learning. It is attractive due to the availability of large
amounts of unlabeled data, and the growing desire to exploit it in
order to improve the quality of predictions and inference in
downstream applications.  Although many proposed methods have been
successful empirically, a formal understanding of the pros and cons of
different semi-supervised methods is still incomplete.

The goal of this paper is to study the tradeoffs between some recently
proposed Laplacian regularization algorithms for graph-based
semi-supervised learning.  In the noiseless setting, the problem
amounts to a particular form of interpolation of a graph-based
function.  More precisely, consider a graph $\graph = (\vertex,
\edge,w)$ where $\vertex =\{v_1,\cdots,v_\numobs\}$ is a set of
$\numobs$ vertices, and $\edge$ is the set of edges equipped with a
set $w = (w_{e})_{e\in \edge}$ of non-negative edge weights.  For some
subset $\ObsSet \subset \vertex$ of the vertex set, say with
cardinality $|\ObsSet |=\numlabeled \ll \numobs$, and an unknown
function $\fstar: \vertex \rightarrow \R$, suppose that we are given
observations $(y_i = f^*(v_i))_{i \in \ObsSet}$ of the function at the
specified subset of vertices. Our goal is to use the observed values
to make predictions of the function values at the remaining vertices
in a way that agrees with $\fstar$ as much as possible.

In order to render this problem well-posed, the behavior of the
function $\fstar$ must be tied to the properties of the graph
$\graph$.  In a statistical context, one such requirement is such that
the marginal distribution of the points $v_i$ be related to the
regression function $\fstar$---for instance, by requiring that
$\fstar$ be smooth on regions of high density.  This assumption and
variants thereof are collectively referred to as the \emph{cluster
  assumption} in the semi-supervised learning literature.

Under such a graph-based smoothness assumption, one reasonable method
for extrapolation is to penalize the change of the function value
between neighboring vertices while agreeing with the observations. A
widely used approach involves using the $\ell_2$-based Laplacian as a
regularizer; doing so leads to the objective
\begin{align}
\label{eq:l2reg}
\min_f \sum_{ij \in \edge} w_{ij} \big(f(v_i) - f(v_j)\big)^2
\quad \mbox{subject to $f(v_i) = y_i$ for all $i \in \ObsSet$,}
\end{align}
where
the penalization is enforced by the quadratic form given by the graph
Laplacian~\citep{zhu2003semi}. This method is closely tied to heat
diffusion on the graph and has a probabilistic interpretation in terms
of a random walk on the graph. Unfortunately, solutions of this
objective are badly behaved in the sense that they tend to be constant
everywhere except for the points $\{v_i\}_{i \in \ObsSet}$ associated
with observations~\citep{nadler2009semi}.  The solution must then have
sharp variations near those points in order to respect the measurement
constraints.

Given this undesirable property, several alternative methods have been
proposed in recent work
(e.g.,~\cite{alamgir2011phase,bridle2013p,zhou2011semi,kyng2015algorithms}).
One such formulation is based on interpolating the observed points
exactly while penalizing the maximal gradient value on neighboring
vertices:
\begin{align}
\min_f \max_{ij\in E} w_{ij} \abs{f(v_i) - f(v_j)} \quad \mbox{subject
  to $f(v_i) = y_i$ for all $i \in \ObsSet$.}
\label{eq:infmin}
\end{align}
Any solution to this variational problem is known as an
\emph{inf-minimizer}.  \cite{kyng2015algorithms} recently proposed a
fast algorithm for solving the optimization
prolblem~\eqref{eq:infmin}. In fact, their algorithm finds a specific
solution that not only minimizes the maximum gradient value, but also
the second largest one among all minimizers of the former and so
on---that is to say, they find a solution such that the gradient
vector $(w_{ij}\abs{f(v_i)-f(v_j)})_{(i,j) \in \edge}$ is minimal in
the lexicographic ordering, which they refer to as the
\emph{lex-minimizer}. They observed empirically that when $|\vertex| =
\numobs$ grows to infinity while both the degree of the graph and
number of observations are held fixed, the lex-minimizer is a better
behaved solution than its $2$-Laplacian counterpart. More precisely,
the observed advantage is twofold: (a) the solution is a better
interpolation of the observed values; and (b) the average
$\ell_1$-error $\frac{1}{n}\sum_{i=1}^n |f(v_i) - \fstar(v_i)|$ for
the lex-minimizer remains stable, while it quickly diverges with
$\numobs$ when $f$ is the $2$-Laplacian minimizer. Their experiments
together with the known limitations of the $2$-Laplacian
regularization method point to the possible superiority of
$\ell_\infty$-based formulation~\eqref{eq:infmin} over the
$\ell_2$-based~\eqref{eq:l2reg} in a semi-supervised setting.
However, we currently lack a theoretical understanding of this
assertion. Accordingly, we aim to fill this gap in the theoretical
understanding of Laplacian-based regularization by studying 
both formulations in the asymptotic limit as the graph size
goes to infinity.

We conduct our investigation in the context of a more general
objective that encompasses the approaches~\eqref{eq:l2reg}
and~\eqref{eq:infmin} as special cases.  In particular, for a positive
integer $p \geq 2$, we consider the variational problem
\begin{equation}
J_p(f) = \sum_{ij\in E} w_{ij}^p \abs{f(v_i) - f(v_j)}^p.
\label{eq:q-energy}
\end{equation}
The objective $J_p$ is referred to as the (discrete) $p$-Laplacian of
the graph $G$ in the literature; for instance, see the papers
by~\cite{zhou2005regularization} and~\cite{buhler2009spectral}, as
well as references therein.  It offers a way to interpolate between
the 2-Laplacian regularization method and the inf-minimization
approach. It is then natural to consider the general family of
interpolation problems based on $p$-Laplacian regularization---namely
\begin{align}
\min_f ~J_p(f) \quad \mbox{subject to $f(v_i) = y_i$ for $i \in
  \ObsSet$.}
\label{eq:qlapreg}
\end{align}
Formulations~\eqref{eq:l2reg} and~\eqref{eq:infmin} are recovered
respectively with $p=2$ and $p \rightarrow \infty$. Indeed, in the
latter case, observe that $\underset{p\rightarrow \infty}{\lim}
J_p(f)^{1/p} = \underset{ij\in E}{\max} ~ w_{ij} \abs{f(v_i) -
  f(v_j)}$. Moreover, under certain regularity assumptions
(\cite{egger1990rate,kyng2015algorithms}), it follows that the
lex-minimizer is the limit of the (unique) minimizers of $J_p$ as $p$
grows to infinity\footnote{The construction of this sequence of
  minimizers is known as \emph{the P\'olya algorithm}, and the study
  of its rate of convergence is a classical problem in approximation
  theory
  (\cite{darst1983polya,egger1987dependence,legg1989polya,egger1990rate}).}---that
is, we have the equivalence
\begin{align}
f_{\text{lex}} = \lim_{p \rightarrow \infty} ~ \underset{u}{\arg \min}
~~ J_p(u) \qquad \mbox{subject to $u(v_i) = y_i$ for all $i \in
  \ObsSet$.}
\end{align}


\paragraph{Our contributions:}

We analyze the behavior of $p$-Laplacian interpolation when the
underlying graph $G$ is drawn from a geometric random model.  Our
first main result is to derive a variational problem that is the
almost-sure limit of the formulation~\eqref{eq:qlapreg} in the
asymptotic regime when the size of the graph grows to infinity.  For a
twice differentiable function $f$, we use $\nabla f$ and $\nabla^2 f$
to denote its gradient and Hessian, respectively.  Letting $\pdens$
denote the density of the vertices in a latent space, we show that for
any even integer $p \geq 2$, solutions of this variational problem
must satisfy the partial differential equation
\begin{align*}
\Delta_2 f(x) + 2\lbr \nabla \log \pdens(x), \nabla f(x)\rbr + (p-2)
\Delta_{\infty} f(x) = 0,
\end{align*}
where $\Delta_2 f \defn \trace(\nabla^2 f)$ is the usual
\mbox{\emph{$2$-Laplacian operator}}, while \mbox{$\Delta_\infty f
  \defn \frac{\lbr \nabla f ,~ \nabla^2f~\nabla f \rbr}{\lbr \nabla f
    , \nabla f \rbr}$} is the \mbox{\emph{$\infty$-Laplacian
    operator,}} which is defined to be zero when $\nabla f = 0$.

This theory then yields several predictions on the behavior of these
regularization methods when the number of labeled examples is fixed
while the number of unlabeled examples becomes infinite: the method
leads to degenerate solutions when $p \le \usedim$; i.e., they are
discontinuous, a manifestation of the curse of dimensionality.  On the other hand, the
solution is continuous when $p \ge d+1$.  The solution \textit{is}
dependent on the underlying distribution of the data for all finite
values of $p$; however, when $p=\infty$, the solution \textit{is not}
dependent on the underlying density $\pdens$. Consequently, as the
graph size increases, the lex- and inf-minimizers end up interpolating
the observed values without exploiting the additional knowledge of the
density $\pdens$ of the features that is provided by the abundance of
unlabeled data.

In order to illustrate the consequences of this last property, we
study a simple one-dimensional regression problem whose intrinsic
difficulty is controlled by a parameter $\epsilon > 0$. We show that
the 2-Laplacian method has an estimation rate independent of
$\epsilon$ while the infinity-minimization approach has a rate that is
increasing in $1/\epsilon$.  As shown by our analysis, this important
difference can be traced back to whether or not the method leverages
the knowledge of $\pdens$. We also provide an array of experiments
that illustrate some pros and cons of each method, and show that our
theory predicts these behaviors accurately.  Overall, our theory lends
support to using intermediate value of $p$ that will lead to
non-degenerate solutions while remaining sensitive to the underlying
data distribution.


\section{Generative Model}

We follow the popular assumption in the semi-supervised learning
literature that the graph represents the metric properties of a cloud
point in $d$-dimensional Euclidean space
(\cite{zhu2003semi,bousquet2003measure,belkin2004semi,hein2006geometrical,nadler2009semi,zhou2011semi}). More
precisely, suppose that we are given a probability distribution on the
unit hypercube $[0,1]^d$ having a smooth density $\pdens$ with respect
to the Lebesgue measure, as well as a bounded decreasing function
$\Gfun :\R_+ \rightarrow \R_+$ such that $\lim_{z \rightarrow \infty}
\Gfun(z)= 0$.  We then draw an i.i.d. sequence $(x_i)_{i=1}^\numobs$
of samples from $\pdens$; these vectors will be identified with the vertices of the
graph $G$: $x_i \equiv v_i$.  Finally, we associate to each pair of vertices the edge weight
$w_{ij} = \Gfun \left( \frac{\elltwo{x_i-x_j}}{\bandwidth} \right)$,
where $\bandwidth > 0$ is a bandwidth parameter.  We use
$\graph_{\numobs, \bandwidth}$ to denote the random graph generated in
this way.


\paragraph{Degree asymptotics}

Given a sequence of graphs $\{ \graph_{\numobs, \bandwidth}
\}_{\numobs=1}^\infty$ generated as above, we study the behavior of
the minimizers of $J_p$ in the limit $\numobs \rightarrow \infty$.  A
first step is to understand the behavior of the graph itself, and in
particular its degree distribution in this limit.  In order to gain
intuition for this issue.  consider the special case when $\Gfun(z) =
\mathbf{1}\{z\le 1\}$.  If the bandwidth parameter $\bandwidth > 0$ is
held fixed, then any sequence $x_{i_1},\cdots,x_{i_k}$ of points that
fall in a ball of radius $h$ will form a clique.  Thus, the graph will
contain roughly $1/h^d$ cliques, each with approximately $\numobs h^d$
vertices. It is typically desired that the sequence of graphs be
sparse with an appropriate degree growth (e.g., constant or
logarithmic growth) so that it converges to the underlying manifold.
In order to enforce this behavior, the bandwidth parameter
$\bandwidth$ should tend to zero as the sample size $\numobs$
increases.

Under this assumption, it can be shown that the \emph{scaled degree}
at any vertex $x$, given by
\begin{align*}
d(x) = \frac{1}{\numobs \bandwidth^d} \sum_{i=1}^\numobs \Gfun \left(
\frac{ \elltwo{x_i-x}}{h} \right),
\end{align*}
concentrates around $\pdens(x)$.  A precise statement can be found
in~\cite{hein2006geometrical}; roughly speaking, it follows from the
fact that for any fixed point $x$ in $[0,1]^d$, as $h$ goes to zero
and under a smoothness assumption on the density $\pdens$, the
probability that a random vector $x_i \sim \pdens$ falls in the
$h$-neighborhood of $x$ scales as $\Pr(\elltwo{x_i-x} \le \bandwidth )
= \int_{\elltwo{z-x} <h} \pdens(z) dz \sim \bandwidth^d \pdens(x)$.


\section{Variational problem and related PDE}

In this section, our main goal is to study the behavior of the
solution in the limit as the sample size $\numobs \rightarrow \infty$
and the bandwidth $\bandwidth \rightarrow 0$. As discussed above, it
is natural to consider scalings under which $\bandwidth$ decreases in
parallel with the increase in $\numobs$.  However, for simplicity, we
follow~\cite{nadler2009semi} and first take the sample size $\numobs$
to infinity with the bandwidth held constant, and then let
$\bandwidth$ go to zero.\footnote{In the case $p=2$, it is known that
  the same limiting objects are recovered when a joint limit in
  $\numobs$ and $\bandwidth$ is taken with $h \rightarrow 0$ but $
  \numobs \bandwidth^d/ \log \numobs \rightarrow \infty$
  (\cite{hein2006geometrical}). Based on the discussion above, this
  scaling implies a super-logarithmic degree sequence. It is still to
  be verified if the same result holds for all $p$.} Our first result
characterizes the asymptotic behavior of the objective $J_p$.

\begin{theorem}
\label{thm:limit}
Let $f$ be continuously differentiable with a bounded derivative, and
let $\pdens$ be a bounded density. Then for any even integer $p \geq 2$, we
have
\begin{align}
I_p(f) \defn \lim_{h \rightarrow 0} ~\lim_{\numobs \rightarrow \infty}
~\frac{1}{\numobs^2 ~h^{p+d}}~ J_p(f) = C_p \int \elltwo{\nabla f
  (x)}^p \pdens^2(x) dx,
\end{align}
where $C_p \defn \frac{1}{d^{p/2}} \int \elltwo{z}^p \Gfun
\big(\elltwo{z} \big)^p dz$.
\end{theorem}

\begin{figure}[t!]
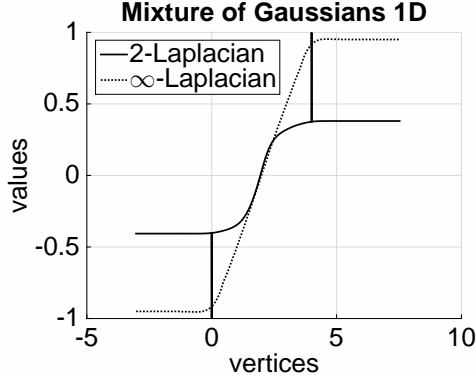

\begin{center}
\widgraph{.4\textwidth}{g1d}
\caption{A mixture of two 1-dimensional Gaussians $N(0,1)$ and
  $N(4,1)$ with equal weights. 500 points are drawn i.i.d.\ from each
  component. We added one point at 0 with label -1 and one point at 4
  with label +1. The similarity graph is constructed with an RBF
  kernel with bandwidth $.4$. }
\label{fig:1d-mix}
\end{center}
\end{figure}
\noindent We provide the proof in Appendix~\ref{sxn:limit}; it
involves applying the strong law for $U$-statistics, as well as an
auxiliary result on the isotropic nature of the integral of a rank-one
tensor. \\

Based on Theorem~\ref{thm:limit}, the asymptotic limit of the
semi-supervised learning problem is a supervised non-parametric
estimation problem with a regularization term given by the functional
$I_p$---namely, the problem
\begin{align}
\label{eq:q-functional}
\underset{g}{\inf} \int \elltwo{\nabla g (x)}^p \, \pdens^2(x) dx
\quad \mbox{subject to $g(x_i) = y_i$ for all $i \in \ObsSet$.}
\end{align}
Our next main result characterizes the solutions of this optimization
problem in terms of a partial differential equation known as the
(weighted) $p$-Laplacian equation.  Here the word ``weighted'' refers
to the term $\pdens^2$ in the functional~\eqref{eq:q-functional}
(\cite{heinonen2012nonlinear,oberman2013finite}).

Let us introduce various pieces of notation that are useful in the
sequel.  Given a vector field $F:\R^\usedim \rightarrow \R^\usedim$,
we use $\DIV(F) \defn \sum_{i=1}^\usedim \partial_{x_i}F_i$ to denote
denote its divergence.  For a scalar-valued function $f:\R^\usedim
\rightarrow \R$, we let
\begin{align*}
\Delta_2 f = \DIV\big(\nabla f \big) = \sum_{i=1}^\usedim
\partial_{x_i}^2f, \quad \mbox{and} \quad \Delta_{\infty} f =
\frac{\lbr \nabla f ,~ \nabla^2f~\nabla f \rbr}{\lbr \nabla f , \nabla
  f \rbr} = \frac{1}{ \elltwo{\nabla f}^2} \sum_{i, j=1}^\usedim
\partial_{x_i} f \cdot \partial_{x_i,x_j} f \cdot \partial_{x_j} f
\end{align*}
denote the (standard) $2$-Laplacian operator and the
$\infty$-Laplacian operator, respectively.

\begin{theorem}
\label{thm:euler_lagrange}
Suppose that the density $\pdens$ is bounded and continuously
differentiable. Then any twice-differentiable minimizer $f$ of the
functional~\eqref{eq:q-functional} must satisfy the Euler-Lagrange
equation
\begin{subequations}
\begin{align}
\DIV \big(\pdens^2(x) \elltwo{\nabla f(x)}^{p-2} \nabla f(x)\big) = 0.
 \label{eq:euler-lagrange}
\end{align}
If moreover the distribution $\pdens$ has full support, then
equation~\eqref{eq:euler-lagrange} is equivalent to
\begin{align}
\Delta_2 f(x) + 2 \lbr \nabla \log \pdens(x), \nabla f(x)\rbr + (p-2)
\Delta_{\infty} f(x) = 0.
\label{eq:q-laplacian}
\end{align}
\end{subequations}
\end{theorem}

The proof employs standard tools from calculus of
variations~\citep{gelfand1963calculus}.  We note here that $f$ does
not need to be twice differentiable for the above result to
hold~(\cite{heinonen2012nonlinear}) in which case
equations~\eqref{eq:euler-lagrange} and~\eqref{eq:q-laplacian} have to
be understood in the viscosity
sense~(\cite{crandall2001optimal,armstrong2010easy}). Twice
differentiability is assumed so only for ease of the proof (see
Appendix~\ref{sxn:euler_lagrange}).

When $\pdens$ is the uniform distribution,
equation~\eqref{eq:q-laplacian} reduces to the partial differential
equation (PDE)
\begin{align*}
\Delta_2 f(x) + (p-2) \Delta_{\infty} f(x) = 0,
\end{align*}
which is known as the $p$-Laplacian equation and often studied in the
PDE literature (\cite{heinonen2012nonlinear,oberman2013finite}).  If
one divides by $p$ and lets $p \rightarrow \infty$, one obtains the
infinity-Laplacian equation
\begin{align}
\label{eq:infinity-laplacian}
\Delta_{\infty} f(x) = 0,
\end{align}
subject to measurement constraints $f(x_i) = y_i$ for $i\in \ObsSet$.
This problem has been studied by various authors
(e.g.,~\cite{crandall2001optimal,aronsson2004tour,peres2009tug,
  armstrong2010easy}). Note that in dimension $d = 1$, we have
$\Delta_{\infty} = \Delta_2 = \frac{d^2}{dx^2}$, and
equation~\eqref{eq:q-laplacian} reduces to
\begin{align*}
(p-1) \pdens(x) f^{''}(x) + 2 \pdens'(x)f'(x)= 0.
\end{align*}
Therefore, if we specialize to the case $p = 2$, the 2-Laplacian
regularization method solves the differential equation $\pdens(x)
f^{''}(x) + 2 \pdens'(x)f'(x)= 0$, whereas if we specialize to $p =
\infty$, then the inf-minimization method solves the differential
equation $f^{''}=0$.  Note that the two equations coincide only when
$\pdens$ is the uniform distribution, in which case $\pdens'$ is
uniformly zero.


\section{Insights and predictions}

Our theory from the previous section allows us to make a number
of predictions about the behavior of different regularization
methods, which we explore in this section.



\subsection{Inf-minimization is insensitive to $\pdens$}

Observe that the effect of the data-generating distribution $\pdens$
has disappeared in equation~\eqref{eq:infinity-laplacian}.  One could
also see this by taking the limit $p \rightarrow \infty$ in the
objective to be minimized in Theorem \ref{thm:limit}---in particular,
assuming that $\pdens$ has full support, we have $I_p(f)^{1/p}
\rightarrow \sup_{[0,1]^d} \|\nabla f(x)\|_2$.

\begin{figure}[t]
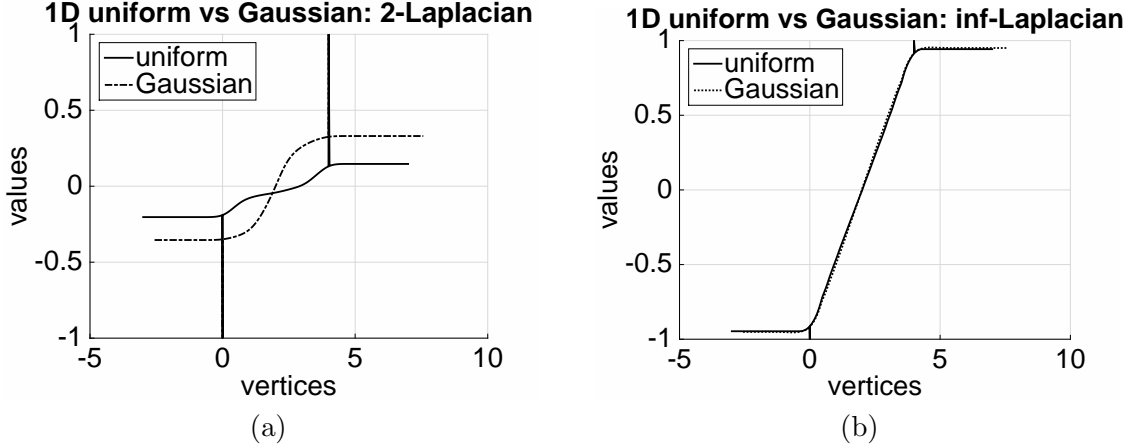

\begin{center}
\begin{tabular}{ccc}
\widgraph{.45\textwidth}{uvg2} &&
\widgraph{.45\textwidth}{uvginf} \\
(a) && (b)
\end{tabular}
\caption{Behavior of the $2$- and infinity- Laplacian solutions under
  a change of input density. The latter is either a mixture of two
  1-dimensional Gaussians $N(0,1)$ and $N(4,1)$ or a mixture of two
  uniform distributions $U([-3,3])$ and $U([1,7])$ with equal
  weights. In each case, 500 points are drawn i.i.d.\ from each
  component. The methods are given two observations $(0,-1)$ and
  $(4,1)$. (a) $\ell_2$-based solution. (b) $\ell_\infty$-based
  solution.}
\label{fig:marginal}
\end{center}
\end{figure}

From the observations above, one can see that in the limit of infinite
unlabeled data---i.e., once the distribution $\pdens$ is
available---the 2-Laplacian regularization method, as well as any
$p$-Laplacian method based on finite (even) $p$, incorporates
knowledge of $\pdens$ in computing the solution; in contrast, for $p =
\infty$, the inf-minimization method does not (see
Figure~\ref{fig:marginal}).  On the other hand, it has been shown that
the $2$-Laplacian method is badly behaved for $d \ge 2$ in the sense
that the solution tends to be uninformative (constant) everywhere
except on the points of observation. The solution must then have sharp
variations on those points in order to respect the measurement
constraints (see Figure~\ref{fig:1d-mix} for an illustration of this
phenomenon). We show in the next section that this problem plagues the
$p$-Laplacian minimization approach as well whenever $p \le d+1$.

\subsection{$p$-Laplacian regularization is degenerate for $p \le d$}
\label{sec:degenq}
In this section, we show that $p$-Laplacian regularization is
degenerate for all $p \leq d$.  This issue was originally addressed
by~\cite{nadler2009semi}, who provided an example that demonstrates
the degeneracy for $p = 2$ for $d \geq 2$.  Here we show that the
underlying idea generalizes to all pairs $(p,d)$ with $p \leq d$.
Recall that Theorem~\ref{thm:limit} guarantees that
\begin{align*}
 I_p(f) & \defn \lim_{h \rightarrow 0} ~\lim_{\numobs \rightarrow
   \infty} ~\frac{1}{\numobs^2~ h^{p+d}}~ J_p(f) = C_p \int \|\nabla f
 (x)\|_2^p \pdens^2(x) dx.
\end{align*}
In the remainder of our analysis, we treat the cases $p \leq d - 1$
and $p = d$ separately.

\paragraph{Case $p \leq d-1$:} Beginning
with the case $p \leq d- 1$, we first set $x_0=0$ and then let $x_1$
be any point on the unit sphere (i.e., $\|x_1\|_2 = 1$).  Define the
function $f_\epsilon (x)= \min\{\|x\|_2/\epsilon,1\}$ for some
$\epsilon \in (0,1)$, and let the observed values be $y_j =
f_\epsilon(x_j)$ for $j \in \{0,1 \}$.  Using the fact that $\nabla
\|x\|_2 = \frac{x}{\|x\|_2}$ and assuming that $\pdens$ is uniformly
upper bounded by $\pmax$ on $[0,1]^d$, we have
\begin{align*}
I_p(f_\epsilon) = \int_{B(0,\epsilon)} \frac{\pdens^2(x)}{\epsilon^p}
dx \leq \frac{\pmaxsq}{\epsilon^p} \vol(B(0,\epsilon)) = \pmaxsq
\vol(B(0,1)) \epsilon^{d-p},
\end{align*}
where $B(0,\epsilon)$ denotes the Euclidean ball of radius $\epsilon$
centered at the origin, and $\vol$ denotes the Lebesgue volume in
$\R^d$.  Consequently, we have $\lim_{\epsilon \to 0}I_p(f_\epsilon) =
0$, so the infimum of $I_p$ is achieved for the trivial function that
is $1$ everywhere except at the origin, where it takes the value
$0$. The key issue here is that $\|\nabla f(x)\|_2^p$ grows at a rate of
$\frac{1}{\epsilon^p}$ while the measure of the ``spike'' in the
gradient shrinks at a rate of $\epsilon^d$.

\paragraph{Case $p = d$:}  On the other hand, when $p = d$,
then we take $f_\epsilon (x)= \log \big( \frac{\|x\|_2^2 +
  \epsilon}{\epsilon} \big) / \log \big( \frac{1 + \epsilon}{\epsilon}
\big)$, for which we also have $y_0 = f_\epsilon(x_0) = 0$ and $y_1 =
f_\epsilon(x_1) = 1$.  With this choice, we have
\begin{align*}
I_p(f_\epsilon) & = \frac{1} { \log \big( \frac{1+\epsilon}{\epsilon}
  \big)^d} \int_{B(0,1)} \frac{\|x\|_2^d}{ \big (
  \|x\|_2^2+\epsilon\big)^d} \pdens^2(x) dx \\
& \leq \frac{\pmaxsq}{\log \big (\frac{1+\epsilon}{\epsilon} \big)^d}
\int_{B(0,1)} \frac{\|x\|_2^d}{ \big(\|x\|_2^2+\epsilon \big)^d} dx\\
& \stackrel{(i)}{=}
\frac{\pmaxsq}{\log\big(\frac{1+\epsilon}{\epsilon}\big)^d} \cdot
\vol(B(0,1)) \int_0^1 \frac{r^d}{\big(r^2+\epsilon\big)^d} dr^{d-1} dr
\\
& \stackrel{(ii)}{=} \frac{d~ \pmaxsq
  \vol(B(0,1))}{\log\big(\frac{1+\epsilon}{\epsilon} \big)^d} \int_0^1
\frac{u^{d-1}}{\big(u+\epsilon\big)^d} du \\
& \stackrel{(iii)}{\leq} \frac{d~ \pmaxsq \vol(B(0,1))}{2 \log \big (
  \frac{1+\epsilon}{\epsilon} \big)^{d-1}},
\end{align*}
where step (i) follows from a change of variables from $x$ to the
radial coordinate $r$; step (ii) follows by the variable change
$u=r^2$; and step (iii) follows by upper-bounding $u$ by $u+\epsilon$
in the numerator inside the integral.  Again, we find that
$\lim_{\epsilon \to 0}I_p(f_\epsilon) = 0$.  Thus, in order to avoid
degeneracies, it is necessary that $p \ge d+1$.


It is worth noting that \cite{alamgir2011phase} studied the problem of
computing the so-called \emph{$q$-resistances} of a graph, which are a
family of distances on the graph having a formulation similar ---in
fact, dual--- to the $p$-Laplacian regularization method considered in
the present paper, and where $1/p +1/q=1$. They established a phase
transition in the value of $q$ for the geometric random graph model,
where above the threshold $q^{**} = 1+1/(d-2)$, the $q$-resistances
``[...] depend on trivial local quantities and do not convey any
useful information [...]" about the graph, while below the threshold
$q^* = 1+1/(d-1)$, these resistances encode interesting properties of
the graph. They conclude by suggesting the use of $p$-Laplacian
regularization with $1/p +1/q^*=1$. The latter condition can be read
$p=d$.  However, as shown by the examples above, this choice is still
problematic, and in fact, the choice $d+1$ is the smallest admissible
value for $p$.

We also note that the example for $d \ge p+1$ extends to an arbitrary
number of labeled points: one simply has a spike for each
point. Undesirable behavior arises as long as the set $\{x_i \mid i
\in \ObsSet\}$ of observed points is of measure zero. Finally, we note
that both the above example can be adapted to the case where the
squared loss $(f(x_i)-y_i)^2$ is optimized along with the regularizer
instead of imposing the hard constraint $f(x_i)=y_i$ (see Appendix~\ref{sxn:least_squares}). 
The issue is that the regularizer is too weak and allows to choose the solution
from a very large class of functions.


\subsection{$p$-Laplacian solution is smooth for $p\ge d+1$}

At this point, a natural question is whether the condition $p\ge d+1$
is also \emph{sufficient} to ensure that the solution is
well-behaved. In a specific setting to be described here, the answer
is \emph{yes}\footnote{Interestingly enough, the $p$-Laplacian
  equation has been extensively studied in non-linear potential
  theory. It is in fact the prototypical example of a non-linear
  degenerate elliptic equation. The regularity of the solutions is
  well understood for any real number $1< p <\infty$ (see
  e.g.\ \cite{heinonen2012nonlinear}). For our purposes however, we do
  not need the full power of this theory.}. The underlying reason is
the Sobolev embedding theorem.  More precisely, let $W^{1,p}([0,1]^d)$
denote the weighted Sobolev space of all (weakly) differentiable
functions $f$ on $[0,1]^d$ such that the semi-norm
\begin{align*}
\|f\|_{1,p} \defn \left(\int \|\nabla f(x)\|_2^p \pdens^2(x)dx\right)^{1/p} <
\infty.
\end{align*}
If, moreover, we assume $\pdens$ is strictly positive almost
everywhere and restrict the above class to functions vanishing on the
boundary, then $\|\cdot\|_{1,p}$ actually defines a norm.  When $p>
d$, and under additional regularity conditions on $\pdens$
(e.g.\ upper- and lower-bounded by constants a.e.), the space
$W^{1,p}$ can be embedded continuously into the space of H\"{o}lder
functions of exponent $1-\frac{d}{p}$, i.e.\ functions $u$ such that
$|u(x)-u(y)| \le c\|x-y\|_2^{1-\frac{d}{p}}$ for all $x,y \in [0,1]^d$
for some dimension-dependent constant $c$.  For details, see Theorem
11.34 of~\cite{leoni2009first} or Lemma 5.17
of~\cite{adams2003sobolev}. ~\cite{brown1992embeddings} provide some
relaxed conditions on $\pdens$. Since the minimizer $f$ of $I_p$ is
such that $I_p(f) = \int \|\nabla f(x)\|_2^p \pdens^2(x) dx< \infty$,
the function $f$ is in the Sobolev class $W^{1,p}$, and therefore it
automatically inherits the H\"older smoothness property, i.e.\ the
$p$-Laplacian solution is smooth for $p \geq d+1$, asymptotically as
$\numobs \to \infty , h\rightarrow 0$\footnote{If the graph is finite,
  then the solution might still contain small spikes as apparent in
  Figures~\ref{fig:1d-mix} and \ref{fig:marginal}.}. Incidentally, via
the examples in the previous section, it is clear that no such
embedding exists if $p \le d$.


\subsection{An example where inf-minimization interpolates well}

By extension to the case $p\rightarrow \infty$, the infinity-Laplacian
solutions also enjoy continuity (this solution is actually Lipschitz based 
on its interpretation as the \emph{absolutely minimal Lipschitz extension} of 
the observations (\cite{aronsson2004tour}). It was also
argued~\cite{kyng2015algorithms}, based on experimental results, that
the inf-minimization method has a better behavior in higher dimensions
in terms of faithfulness to the observations.  We illustrate this
point by considering a simple example, similar to the one above, for
which the $\infty$-Laplacian equation~\eqref{eq:infinity-laplacian}
produces a sensible solution. With $x_0 = 0$ and $S \defn \{x \in \R^d
\mid \|x\|_2 = 1\}$ denoting the Euclidean unit sphere, suppose that we
limit ourselves to functions satisfying the observation constraints
$y_0 = f(0) =0$ and $y(x) = 1$ for all $x \in S$.

Without any further information on the data-generating process, a
reasonable fit is the function $\fbar(x) = \|x\|_2$. We claim that it
is the only radially symmetric solution to the differential equation
$\Delta_{\infty} f = 0$ with the boundary constraints $f(x_0)=y_0$ and
$f(x)=y(x)$ for all $x \in S$.  In order to verify this claim, let
$f(x) = g(\|x\|_2)$ for a function $g: \R_+\rightarrow \R$. For any
non-zero $x \in \real^d$, we have
\begin{align*}
\nabla f(x) = g'(\|x\|_2)\frac{x}{\|x\|_2}, \quad \text{and} \quad
\nabla^2f(x) = g''(\|x\|_2) \frac{xx^\intercal}{\|x\|_2^2} +
g'(\|x\|_2)\frac{I}{\|x\|_2} -g'(\|x\|_2)
\frac{xx^\intercal}{\|x\|_2^3}.
\end{align*}
Then
\begin{align*}
\Delta_\infty f(x) = \frac{1}{\|\nabla f(x)\|_2^2}\nabla f(x)^\intercal
\nabla^2 f(x) \nabla f(x) = g''(\|x\|_2).
\end{align*}

Given the boundary conditions on $f$, the only solution to
$\Delta_\infty f$ = 0, is given by $g(r) = r$, meaning that $f(x) =
\|x\|_2$. On the other hand, the latter is not a solution to $\Delta_2
f=0$, unless $d=1$.

\vspace{.5cm}

In summary, this section reveals a trade-off between smoothness and
sensitivity to the data-generating density $\pdens$ in $p$-Laplacian
regularization: the solution is strongly sensitive to $\pdens$ but is
non-smooth for small values of $p$, while it is smooth but weakly
dependent on $\pdens$ for large and infinite values of $p$. The
transition from degeneracy to smoothness happens at a sharp threshold
$p^*=d+1$, while the dependence on $\pdens$ weakens with larger and
larger $p$ without a threshold.

While the property of smoothness is an obvious quality in an
estimation setting, and may lead to improved statistical rates if
assumed first hand ---especially if the signal one wishes to recover
is itself smooth--- it is less obvious how to quantify the advantages
entailed by the sensitivity to the underlying data-generating density;
especially when the latter is available and is to be incorporated in
the design of an estimator. We provide in the next section a simple,
one-dimensional regression example where the regression function is
tied to the density $\pdens$ via the cluster assumption, and where a
difference in estimation rates between the $\ell_2$ and $\ell_\infty$
methods is exhibited. This difference is explicitly due to the fact
that the $\ell_2$ method leverages the knowledge of $\pdens$ while
$\ell_\infty$ does not.


\section{The price of ``forgetting'' $\pdens$}

We consider in this section a simple estimation example in one
dimension where the asymptotic formulation of 2-Laplacian
regularization method achieves a better rate of convergence than that
of the inf-minimization method. Such an advantage of the $\ell_2$
method over the $\ell_\infty$ method should be conceivable under the
cluster assumption: the regularizer $I_{2} = \int f'^2 \pdens^2$ will
encodes information about the target function via $\pdens$ while the
regularizer $I_\infty = \sup |f'|$ does not.

Let the target function $\fstar$ and the data-generating density
$\pdens$ be supported on the interval $[-1,1]$. For some small
$\epsilon > 0$, we construct a density $\pdens$ that takes a uniformly
small value over the interval $[-\epsilon,\epsilon]$, and takes and a
large value on the complementary set $[-1,-\epsilon) \cup
  (\epsilon,1]$. We also let $\fstar$ have a high Lipschitz constant
on the interval $[-\epsilon,\epsilon]$ and be constant otherwise.
More precisely, the density $\pdens$ and function $\fstar$ are
constructed as follows:
\begin{equation}
\begin{array}{cc}
\hspace{-1cm} \pdens(x) = \begin{cases} b & x \in
  [-\epsilon,\epsilon],\\ a & x \in [-1,-\epsilon)\cup(\epsilon,1],
\end{cases}
~~ & ~~ \fstar(x) =
\begin{cases}
-1 & x \in [-1,-\epsilon),\\ x/\epsilon & x \in
  [-\epsilon,\epsilon],\\ 1 & x \in (\epsilon,1].
\end{cases}
\end{array}
\end{equation}

The constants $a$ and $b$ are related by the equation $(1-\epsilon)a +
\epsilon b = 1/2$ so that the density $\pdens$ integrates to 1, and we
think of $b$ as being much smaller than $a$, i.e. $b \ll a$.  Consider the
following two classes of functions corresponding to the regularizer
$I_p$ for $p=2$ and $p=\infty$ respectively:
\begin{align*}
\Hil & \defn \Big\{ f : [-1,1] \rightarrow \R ~,~ f ~\text{absolutely
  continuous, odd and} ~ \int_{-1}^1 f'(x)^2 \pdens^2(x) dx < \infty
\Big\},\\
\Lfun & \defn \Big\{ f : [-1,1] \rightarrow \R ~,~ f ~\text{absolutely
  continuous, odd and} ~\underset{|x| \le 1}{\sup} |f'(x)| < \infty
\Big\}.
\end{align*}

We define the associated norms $\|f\|_{\Hil} \defn \big(\int_{-1}^1
f'(x)^2 \pdens^2(x) dx \big)^{1/2}$ and $\|f\|_{\Lfun} \defn
\underset{x \in [-1,1]}{\sup} |f'(x)|$ on $\Hil$ and $\Lfun$
respectively.  Observe that $\|f^*\|_{\Lfun} = 1/\epsilon$ while
$\|f^*\|_{\Hil}$ is upper bounded by a constant:
\begin{align*}
\int_{-1}^1 [(f^{*})'(x)]^2 \pdens^2(x) dx = \int_{-\epsilon}^\epsilon
\frac{1}{\epsilon^2} b^2 dx = 2b^2/\epsilon.
\end{align*}
Taking $b = \sqrt{\epsilon}$, the above integral is bounded above by
$2$.

We draw ${\numlabeled}$ points $(x_i)_{i=1}^{{\numlabeled}}$
independently from $\pdens$ and observe the responses $y_i =
\fstar(x_i) + \sigma^2 \xi_i$ where $\xi_i \sim N(0,1)$ are
i.i.d.\ standard normal random variables, $\sigma >0$. We compare the
following two M-estimators:
\begin{center}
\begin{tabular}{ccc}
$\begin{aligned} \fhat_\Hil = \arg\min_{f} &~\frac{1}{{\numlabeled}}
    \sum_{i=1}^{\numlabeled} \big(y_i - f(x_i)\big)^2 \\ \text{s.t.\ }
    & ~ f \in \Hil ~,~ \|f\|_{\Hil} \le 2,
\end{aligned}$
\hspace{.5cm}
&
~~ \text{and} ~~
&
\hspace{.5cm}
$\begin{aligned}
\fhat_\Lfun = \arg\min_{f} &~\frac{1}{{\numlabeled}} \sum_{i=1}^{\numlabeled} \big(y_i - f(x_i)\big)^2 \\
\text{s.t.\ } & ~ f \in \Lfun ~,~ \|f\|_{\Lfun} \le 1/\epsilon,
\end{aligned}$

\end{tabular}
\end{center}
in terms of the rate of decay of the error $\big\|
\fhat-\fstar\big\|_{\numlabeled}^2$, where $\|f\|_{\numlabeled}^2 =
\frac{1}{\numlabeled}\sum_{i=1}^{\numlabeled} f^2(x_i)$. Note that
this error can in principle tend to zero as $\numlabeled$ grows to
infinity since the target $\fstar$ belongs to the hypothesis class in
both of the considered cases, i.e.\ there is no approximation error.

%

\begin{theorem}
There are universal constants $(c_0,
c_1, c_2, c_3)$ such that for any $\epsilon \in (0,1/2)$, 
the $\ell_2$ estimator satisfies the bound
\begin{align}
\left\|\hat{f}_\mathcal{H}-\fstar\right\|_{\numlabeled}^2 & \leq c_1
\, \left(\frac{\sigma^2}{\numlabeled}\right)^{2/3}
\label{eq:rate_l2}
\end{align}
with probability at least $1 - \exp \big\{ { -c_0 \big(
  \frac{{\numlabeled}}{\sigma^2} \big)^{1/3}} \big\}$. On the other
hand, the $\ell_\infty$ estimator satisfies the bound
\begin{align}
\big\| \fhat_\Lfun-\fstar\big\|_{{\numlabeled}}^2 & \leq c_3
\Big(\frac{\sigma^2}{\epsilon~ {\numlabeled}}\Big)^{2/3}
\label{eq:rate_inf}
\end{align}
with probability at least $1- \exp \big\{ {-c_2 \big(\epsilon^2
  \frac{{\numlabeled}}{\sigma^2}\big)^{1/3}} \big\}$.
\label{thm:1d_rates}
\end{theorem}

\begin{figure}[!t]
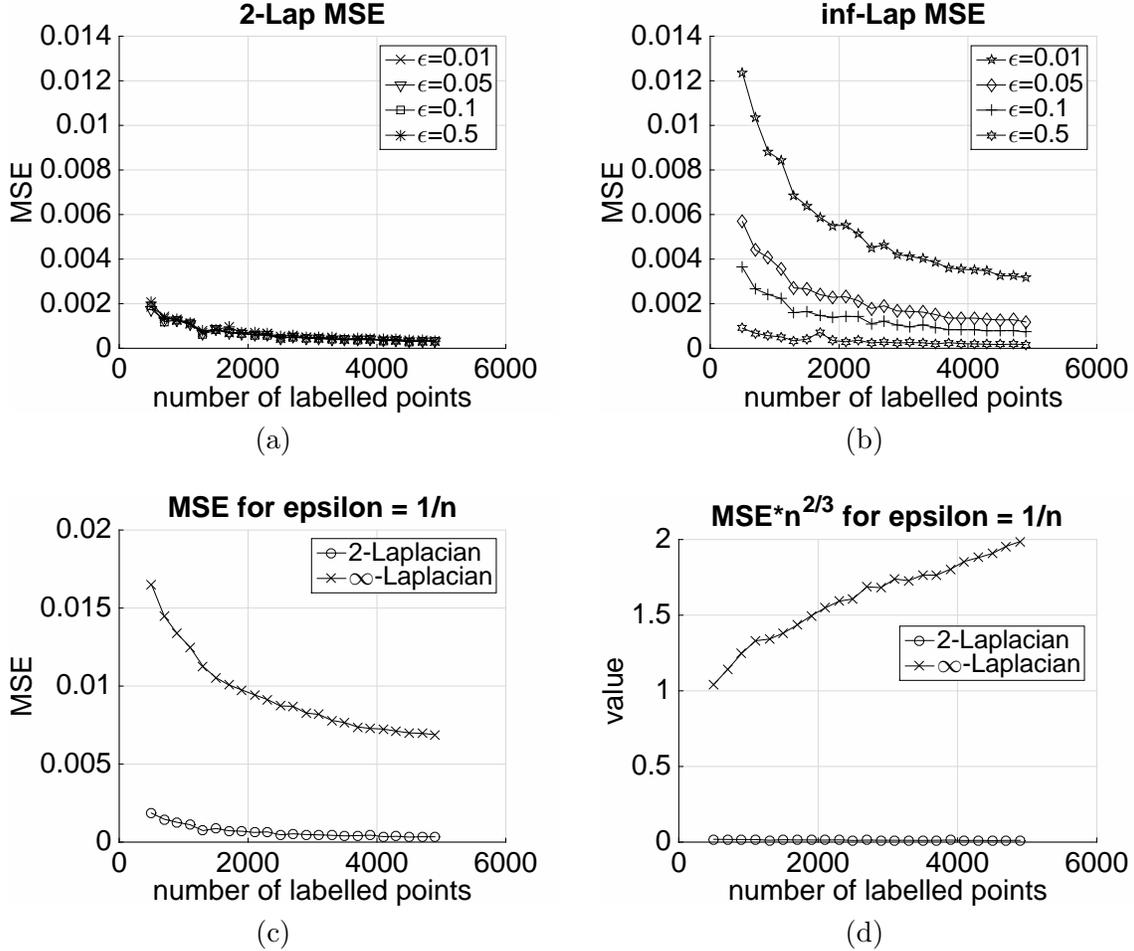

\begin{center}
\begin{tabular}{ccc}
\widgraph{.45\textwidth}{cst_cross_2} &&
\widgraph{.45\textwidth}{cst_cross_inf} \\
(a) & & (b) \\
& & \\
\widgraph{.45\textwidth}{1byn} &&
\widgraph{.45\textwidth}{r1byn_times_n23} \\
(c) & & (d)
\end{tabular}
\caption{ Plots of mean-squared-error (MSE) against number of labeled
  samples.  The regularization parameter is determined using
  cross-validation (thereby producing estimators of \emph{superior}
  performance than $\fhat_{\Hil}$ and $\fhat_{\Lfun}$).  The samples
  $x_i$ are drawn according to $\pdens$ and $y_i = \fstar(x_i) +
  \xi_i$, and $\xi_i$ are i.i.d.\ $N(0,0.05)$.  Panels (a) and (b):
  MSE of $\ell_2$ and $\ell_\infty$ methods respectively for various
  values of $\epsilon$.  As expected, the MSE of the $\ell_2$ method
  is independent of $\epsilon$ while that of the $\ell_\infty$ method
  is increasing in $1/\epsilon$.  Panels (c) and (d): Plots of the MSE
  and MSE~$\times ~{\numlabeled}^{2/3}$, respectively versus the
  sample sample size ${\numlabeled}$ for both methods.  Both plots
  correspond to sequences of problems with $\epsilon =
  1/{\numlabeled}$; panel (d) shows that in this regime, the rate of
  the $\ell_2$ method is roughly ${\numlabeled}^{-2/3}$, thereby
  providing evidence that the upper bound~\eqref{eq:rate_l2} is tight,
  while the rate of the $\ell_\infty$ method, call it $r(\epsilon,
  {\numlabeled})$, is such that $ r(\frac{1}{{\numlabeled}},
  {\numlabeled}) \gg {\numlabeled}^{-2/3}$.  }
\label{fig:MSE}
\end{center}
\end{figure}

We can thus compare upper bounds on the rate of estimation of the
$\ell_2$ and $\ell_\infty$ methods respectively. The upper
bound~\eqref{eq:rate_inf} shows a dependence in $1/\epsilon^{2/3}$,
 while the bound~\eqref{eq:rate_l2} shows no dependence on $\epsilon$. 
 One might ask if these bounds are tight.  In particular, a
question of interest is whether the $\ell_\infty$ estimator
\emph{adapts} to the target $\fstar$ without the need to know the
density $\pdens$, in which case the corresponding estimator would achieve a
better rate. While we think our bounds could be sharpened, 
we strongly suspect that the $\ell_\infty$ estimator cannot achieve a rate independent 
of $\epsilon$ (in contrast to the $\ell_2$ estimator). 
We provide an array of simulations showing that the rate of $\ell_\infty$
deteriorates as $\epsilon$ gets small, where the rate of the $\ell_2$
method stays the same (see Figure~\ref{fig:MSE}).

\subsection{Main ideas of the proof}

The first non-asymptotic bound~\eqref{eq:rate_inf} in
Theorem~\ref{thm:1d_rates} follows in a straightforward way from known
results on the minimax rate of estimation on the class of Lipschitz
functions. Indeed, the rate of estimation on this class with Lipschitz
constant $L$ is $\big(L\sigma^2/{\numlabeled}\big)^{2/3}$, and in our
case $L = 1/\epsilon$.  On the other hand, the second
result~\eqref{eq:rate_l2} follows by recognizing that the class $\Hil$
is a weighted Sobolev space of order 1, which is a Reproducing Kernel
Hilbert Space (RKHS).  The associated kernel, as identified
by~\cite{nadler2009semi}, is given by
\begin{align}
\Ker(x,y) = \frac{1}{4} \int_{-1}^1 dt/ \pdens^2(t) - \frac{1}{2} \left
| \int_{x}^y dt/ \pdens^2(t)\right|, \quad \mbox{for all $x,y \in
  [-1,1]$.}
\label{eq:kernel}
\end{align}

It is known that the rate of estimation on a ball of radius $R$ of an
RKHS is tightly related to the decay of the eigenvalues of the
kernel. More precisely, the rate of estimation is upper-bounded with
high probability by the smallest solution $\delta >0$ to the
inequality
\begin{align}
 \label{eq:master_eq}
 \left(\frac{2}{{\numlabeled}} \sum_{j=0}^\infty
 \min\big\{\kereig_j,\delta^2\big\}\right)^{1/2} \le
 \frac{R}{\sigma}~\delta^2,
\end{align}
where $(\kereig_j)_{j\ge 0}$ is the sequence of eigenvalues of the
kernel
$\Ker$~(\cite{koltchinskii2006local,mendelson2002geometric,bartlett2002localized,van2000empirical}).
Our next result upper-bounds the rate of decay of these eigenvalues.

\begin{lemma}
For any $\epsilon \in (0, 1/2)$, the eigenvalues of the kernel $\Ker$
form a doubly indexed sequence $(\kereig_{\kind, \jind})$ with ${0 \le
  \kind \le 2\kind_0-1, \jind \ge 0}$, $\kind_0 =
\lfloor\sqrt{2}\epsilon^{-3/4}\rfloor$. This sequence satisfies the
upper bound
\begin{align*}
\kereig_{\kind, \jind} & \leq \begin{cases} 1.26 & \mbox{if $\kind =
    \jind = 0$} \\
\frac{1}{\big(\frac{k}{2\sqrt{2}}+j\epsilon^{-3/4}\big)^{2}\pi^2} &
\mbox{otherwise.}
\end{cases}
\end{align*}
\label{lem:eigenvalues}
\end{lemma}
Plugging these estimates in equation~\eqref{eq:master_eq} leads to the
rates we claim in Theorem~\ref{thm:1d_rates}.  The full details are
given in Appendix~\ref{sxn:rates_proof}.


\section{Related work}

The discrete graph $p$-Laplacian was introduced
by~\cite{zhou2005regularization} as a form of regularization
generalizing classical Laplacian regularization in semi-supervised
learning~\cite{zhu2003semi}. We mention however that the continuous
$p$-Lapalcian has been extensively studied much earlier in PDE
theory~\cite{heinonen2012nonlinear,aronsson2004tour}.  It was also
used and analyzed for spectral clustering, where it provides a family
of relaxations to the normalized cut
problem~\cite{amghibech2003eigenvalues,buhler2009spectral,luo2010eigenvectors}.
A dual version of the problem (\emph{the $q$-resistances} or
\emph{$q$-volatges} problem) was investigated and shown to yield
improved classification performance
in~\cite{bridle2013p}. \cite{alamgir2011phase} prove the existence of
a phase transition under the geometric graph model roughly similar to
the one exhibited in this paper, although the thresholds are slightly
different. The exact nature of the connection is still unclear
however. On the other hand, a game-theoretic interpretation of the $p$-Laplacian solution
is studied in~\cite{peres2008tug,peres2009tug}, and a similar transition at $p=d+1$ in the behavior of the game is found.
 The assumption that the graph entails a geometric structure
is popular in the analysis of semi-supervised learning
algorithms~\cite{belkin2001laplacian,bousquet2003measure,belkin2004semi,hein2006geometrical,hein2006uniform,nadler2009semi}. This
line of work have mostly focused on the $2$-Laplacian formulation and
its convergence properties to a differential operator on the limiting
manifold.

Among other approaches that circumvent the degeneracy issue discussed
in the paper, we mention higher order
regularization~\cite{zhou2011semi}, where instead of only penalizing
the first derivative of the function, one can penalize up to $l$
derivatives.  This approach considers solutions in a higher order
Sobolev space $W^{l,2}$ which, via the Sobolev embedding theorem, only
contains smooth functions if $l > d/2$ (see
\cite{adams2003sobolev,leoni2009first}).  This approach can be
implemented algorithmically using the discrete \emph{iterated
  Laplacian}~\cite{zhou2011semi,wang2014trend}.

Results on statistical rates for semi-supervised learning problems are
very sparse. The first results are covered in
\cite{castelli1996relative} in the context of mixture models,
\cite{rigollet2006generalization} in the context of classification,
and \cite{lafferty2007statistical} for regression.  A recent line of
work considers the setting where the graph is fixed while the set of
vertices where the labels are available is
random~\cite{ando2007learning,johnson2007effectiveness,johnson2008graph,rakesh2014learning,shivanna2015spectral}. The
methods studied in this setting generalize the 2-Laplacian method by
penalizing by the quadratic form given by a general positive
semidefinite kernel.  The derived rates depend on the structural
properties of the graph and/or the kernel used. It is shown in
particular that using the normalized Laplacian instead of the regular
one leads to a better statistical bound, and on the other hand, one
can obtain rates depending on the \emph{Lov\'{a}sz Theta function} of
the graph by choosing the regularization kernel optimally.


\section{Conclusion and open problems}

In this paper, we used techniques and ideas from PDE theory to yield
insight into the behavior of various methods for semi-supervised
learning on graphs.  The $d$-dimensional geometric random graph model
analyzed in this paper is a common one in the literature, and the most
common Laplacian penalization technique is for $p=2$, though the
choice $p=\infty$ has also attracted some recent attention.  Our paper
sheds light on both of these options, as well as the full range of $p$
in between. From our asymptotic analysis, we see that for a
$d$-dimensional problem, degenerate solutions occur whenever $p \le
d$, whereas at the other extreme, the choice $p = \infty$ leads to
solutions that are totally insensitive to the input data distribution.
Hence, the choice $p=d+1$ seems like a prudent one, trading off
degeneracy of the solution with sensitivity to the unlabeled data.

An important companion problem is the unconstrained version of the
problem, in which we penalize a (weighted) sum of two terms, the
$p$-Laplacian term and a sum of squared losses on the labeled data.
One can see that under our asymptotics, we can make the same
conclusions about the unconstrained solution. Hence, the conclusions
of this paper do not hinge on the fact that we modeled the problem
with equality constraints, and do apply more generally.  For
completeness, we outline this argument in
Appendix~\ref{sxn:least_squares}.

There are perhaps two important assumptions which must be relaxed in
future work. The first is the asymptotics we consider, in which the
number of labeled points is fixed as the unlabeled points become
infinite.  An interesting situation is when both labeled and unlabeled
points grow at a relative rate. We showed that in the first situation,
a certain class of methods, namely all $p$-Laplacian methods with $p
\le d$, behave poorly.  

An interesting direction is to understand what set of methods are
appropriate in different regimes of relative growth rate. In
particular, we suspect that most of our results should continue to
hold as long as the number of unlabeled points grow at a much faster
rate than the number of labeled points.  The second assumption is
about the geometric random graph model, and how much our results are
tied to this model.  Finite sample rates and non-asymptotic arguments
are necessary to understand how soon we can expect to see these
effects on general graphs in practice.  Finally, the model selection
problem of what $p$ to use in practice is very important, since we may
not know the underlying dimensionality of the data from which our
graph was formed.

\paragraph{Acknowledgments:}{We thank Jason Lee, Kevin Jamieson, Anup Rao, Sushant Sachdeva and Ryan Tibshirani for
  discussions in the early phases of this work. This
  work was partially supported by Air Force Office of Scientific
  Research grant FA9550-14-1-0016 and Office of Naval Research grant
  DOD-ONR-N00014 to MJW. This material is based upon work supported in part by the Office of
Naval Research under grant number N00014-11-1-0688 and by the Army Research
Laboratory and the U. S. Army Research Office under grant number W911NF-11-1-0391to MIJ.}

\bibliographystyle{alpha}
\bibliography{lipschitz}

\appendix

\section{Proof of Theorem~\ref{thm:limit}}
\label{sxn:limit}

Let $x$ and $x'$ be i.i.d. draws from $\pdens$.  Since $f$ is a
bounded function, the first moment $\E\big[|f(x)-f(x')|^q\big]$ is
finite. Therefore, by the strong law of large numbers for
U-statistics~(\cite{serfling2009approximation}), we have
\begin{align*}
\lim_{\numobs \rightarrow \infty} \frac{1}{\numobs^2} J_p(f) =
\int\int \Gfun \left(\frac{\|x-x'\|_2}{h}\right)^p \abs{f(x) - f(x')}^p
\pdens(x) \pdens(x') dx dx', \quad \mbox{almost surely.}
\end{align*}
Writing $x' = x + h z$ for some scalar $h > 0$ and vector $z$, the
second integral simplifies to
\begin{align*}
\int \Gfun \left(\frac{\|x-x'\|}{h}\right)^p \abs{f(x) - f(x')}^p \pdens
(x')dx' = h^d \int \Gfun \big(\elltwo{z}\big)^p \abs{f(x) - f(x+h
  z)}^p \pdens(x+h z) dz.
\end{align*}
We now divide by $\bandwidth^{d+p}$ and consider the behavior as the
bandwidth parameter $\bandwidth$ tends to zero. Since the functions
$f$, $f'$ and $\pdens$ are all bounded on a compact domain, the dominated
convergence theorem implies that
\begin{align*}
\pdens(x) \int \Gfun \big(\elltwo{z}\big)^p \abs{\lbr \nabla f(x) , z
  \rbr}^p dz = \pdens(x) \big \lbr \nabla f(x)^{\otimes p} ,\int \Gfun
\big(\elltwo{z}\big)^p z^{\otimes p} dz \big \rbr.
\end{align*}
Note the above inner product involves tensors of order $p$, and recall
that we have assumed that $p \geq 2$ is even. Since the function
$\Gfun$ depends only on the norm of $z$ in the above integral, the
latter should also be isotropic.  The precise statement is as follows:
\begin{lemma}
\label{lem:tensorlemma}
For any function $w: \R_+ \rightarrow \R_+$ and vector $u \in \R^d$,
we have
\begin{align}
\label{tensor}
 \big \lbr u^{\otimes p}~,~ \int w\big(\elltwo{z}\big) z^{\otimes p}
 dz \big\rbr & = \begin{cases} \frac{1}{d^{p/2}}\big(\int
   w\big(\elltwo{z} \big) \elltwo{z}^{p} dz\big) \cdot \elltwo{u}^p &
   \mbox{if $p \geq 2$ is even} \\
0 & \mbox{if $p$ is odd.}
 \end{cases}
\end{align}
\end{lemma}
Applying Lemma~\ref{lem:tensorlemma} with the function $w(\elltwo{z})
\defn \Gfun (\elltwo{z})^p$ and then simplifying yields
 \begin{align*}
\lim_{\numobs\rightarrow \infty} \frac{1}{\numobs^2 h^{p+d}} J_p(f) =
\frac{1}{d^{p/2}}\int \elltwo{z}^{p} \Gfun \big(\elltwo{z}\big)^p dz
\cdot \int \|\nabla f (x)\|_2^p \pdens^2(x) dx,
\end{align*}
which concludes the proof of the theorem. \\

The only remaining detail is to prove Lemma~\ref{lem:tensorlemma}.


\paragraph{Proof of Lemma~\ref{lem:tensorlemma}}

We proceed by induction on the integer $p$. In the base case ($p=2$),
we have
\begin{align*}
\big \lbr u u^\intercal~,~ \int w\big(\elltwo{z} \big) zz^\intercal dz
\big \rbr = u^\intercal \big(\int w\big(\elltwo{z} \big) zz^\intercal
dz \big)u.
\end{align*}
Since the function $w$ depends only on $\elltwo{z}$, the matrix
between the parentheses above is proportional to the identity, the
proportionality constant can be determined by taking a trace. We end
up with $\int w\big( \elltwo{z} \big) z z^\intercal dz =
\frac{1}{d}\big(\int w\big(\elltwo{z} \big) \elltwo{z}^2 dz\big)
I$. This establishes the base case.\footnote{The case $p=2$ is also
  proven in Proposition 4.1 of the paper~\citep{bousquet2003measure}}
Now assume that for a given even $p \geq 2$, and for all function
non-negative maps $w$, one has \eqref{tensor}. We prove that the same
is true for $p+2$. Define $T_{p} \defn \int w\big(\elltwo{z}\big)
z^{\otimes p} dz$, and for any vector $u \in \R^{d}$, let $\bar{T}_p$
be the partial contraction of $T_{p+2}$ by $u \otimes u$, namely
\begin{align*}
\bar{T}_p \defn & T_{p+2} (u \otimes u) = \int w \big( \elltwo{z}
\big) z^{\otimes p} \lbr u, z \rbr^2 dz.
\end{align*}
The tensor $\bar{T}$ is of order $p$, and the map $z \rightarrow
w(\elltwo{z})\lbr u, z \rbr^2$ is non-negative, so by the induction
hypothesis, for every $v \in \R^d$, we have
\begin{align*}
\lbr v^{\otimes p}~,~\bar{T}_p \rbr = \frac{1}{d^{p/2}} \left ( \int w
\left(\elltwo{z} \right) \elltwo{z}^p \lbr u, z \rbr^2 dz \right)
\elltwo{v}^p.
\end{align*}
By recourse to the base case, the quadratic form between the
parentheses is equal to \\ $\frac{1}{d} \left (\int w \big(\elltwo{z}
\big) \elltwo{z} ^{p+2} dz\right) \elltwo{u} ^2$. Taking $u=v$ completes
the proof of the lemma.


\section{Proof of Theorem~\ref{thm:euler_lagrange}}
\label{sxn:euler_lagrange}

Recall the shorthand notation $I_p(f) \defn \int \elltwo{\nabla f
  (x)}^p \pdens^2(x) dx$. By convexity, the function $f$ is a
minimizer of the functional $I_p$ if for all test functions $h$ and
all sufficiently small real numbers $\epsilon >0$, we have
$I_p(f+\epsilon h) \ge I_p(f)$. Moreover, by a Taylor series
expansion, we have
\begin{align*}
I_p(f+\epsilon h) = I_p(f)+ p\epsilon \int \lbr \nabla f(x), \nabla
h(x)\rbr \cdot \elltwo{\nabla f(x)}^{p-2} \pdens^2(x) dx +
\order(\epsilon^2),
\end{align*}
where the $\order(\epsilon^2)$ term is non-negative by convexity of
$I_p$. Hence, the function $f$ is a minimizer if and only if
\begin{align*}
\int \lbr\nabla f(x), \nabla h(x)\rbr \cdot \elltwo{\nabla f(x)}^{p-2}
\pdens^2(x) dx = 0
\end{align*}
for all testing functions $h$. By integrating by parts and choosing
$h$ to vanish on the boundary of the set $[0,1]^d$, we find that the
above quantity is equal to
\begin{align*}
\int \lbr\nabla f(x), \nabla h(x) \rbr \cdot \elltwo{\nabla
  f(x)}^{p-2} \pdens^2(x) dx = - \int \DIV \big (\pdens^2(x) \elltwo{\nabla
  f(x)}^{p-2} \nabla f(x) \big) h(x) dx.
\end{align*}
This expression has to vanish for all test functions $h$ (that vanish
on the boundary, which implies the Euler-Lagrange equation
\begin{align*}
\DIV \big (\pdens^2(x) \elltwo{\nabla f(x)}^{p-2} \nabla f(x) \big) = 0.
\end{align*}

We now further manipulate this equation so as to obtain the
$p$-Laplacian equation.  In particular, some straightforward
computations yield
\begin{align*}
\partial_{x_i} \big(\pdens^2 \elltwo{\nabla f}^{p-2} \partial_{x_i}
f\big)(x) & = \partial_{x_i}\big ( \pdens^2(x) \elltwo{\nabla
  f(x)}^{p-2}\big)\partial_{x_i}f(x) + \pdens^2(x) \elltwo{\nabla
  f(x)}^{p-2}\partial_{x_i}^2 f(x), \quad \mbox{and} \\
\partial_{x_i}\big(\pdens^2\|\nabla f\|^{p-2}\big)(x) & =
2\partial_{x_i}\pdens(x)\cdot \pdens(x) \|\nabla f(x)\|^{p-2} + \pdens^2(x)
(p-2)\big(\sum_{j=1}^d \partial_{x_i,x_j} f \partial_{x_j}f\big)\cdot
\|\nabla f(x)\|^{p-4}.
\end{align*}
Now summing these terms yield
\begin{align*}
 \DIV \big(\pdens^2(x) \elltwo{\nabla f(x)}^{p-2} \nabla f(x) \big) & = 2
 \pdens(x) \elltwo{\nabla f(x)}^{p-2} \lbr \nabla \pdens(x), \nabla f(x)\rbr +
 \pdens^2(x) \elltwo{\nabla f(x)}^{p-2} \Delta_2 f(x) \\ & + (p-2) \pdens^2(x)
 \elltwo{\nabla f(x)}^{p-4} \big(\sum_{i, j =1}^d \partial_{x_i}f
 \cdot \partial_{x_i,x_j} f \cdot \partial_{x_j}f\big)(x)\\
&= \pdens^2(x) \elltwo{\nabla f(x)}^{p-2} \cdot \big(\Delta_2 f(x) +
 \frac{2}{\pdens(x)} \lbr \nabla \pdens(x), \nabla f(x)\rbr + (p-2)
 \Delta_{\infty} f(x)\big).
 \end{align*}
From the derivation above, the Euler-Lagrange
equation~\eqref{eq:euler-lagrange} is equivalent to
\begin{align*}
\Delta_2 f(x) + 2\lbr \nabla \log \pdens(x), \nabla f(x)\rbr + (p-2)
\Delta_{\infty} f(x) = 0,
\end{align*}
as claimed.

\section{Proof of Theorem~\ref{thm:1d_rates}} \label{sxn:rates_proof}

Bounding the error of $M$-estimators is a classical problem in
statistics and learning theory.  Optimal rates typically follow by
deriving uniform convergence bounds over a small ball \emph{localized}
around the true regression function $\fstar$; for instance, see the
book~\cite{van2000empirical} as well as the
papers~\cite{koltchinskii2006local,bartlett2002localized}). Uniform
convergence is established by upper-bounding the Rademacher or
Gaussian complexity of this small ball via generic covering number
arguments, or by leveraging the special structure of the ball. The
first approach will be used to analyze the rate of the estimator
$\fhat_{\Lfun}$, and the second approach to analyze the rate of
$\fhat_{\Hil}$. In this latter case, the analysis is based on
the study of the spectrum of a certain integral operator associated to
the kernel $k$~\eqref{eq:kernel} that generates the space $\Hil$.

\subsection{Proof of the error bound~\eqref{eq:rate_inf} on $\fhat_{\Lfun}$}

For a given metric space $(\Fclass, \rho)$, we let $N(t, \Fclass, \rho)$
be the covering number of a metric space $\Fclass$ in the metric
$\rho$ at resolution $t$.
Now consider the shifted function class
\begin{align*}
\Lfun^* &  \: \defn \; \{f-\fstar \, \mid \, f\in
 \Lfun ~,~ \|f\|_{\Lfun} \le 1/\epsilon\}
\end{align*}
under the metric $\|f\|_\Lfun = \sup_{x} |f'(x)|$.
By known results on metric entropy~\citep{dud99}, we have $\log(N (t; \Lfun^*;
\|\cdot\|_\Lfun)) = \order \big(\frac{1}{\epsilon t}\big)$, using the
fact that any function in $\Lfun^*$ must be $2/\epsilon$-Lipschitz.

Now let $\|f\|_\numlabeled ^2 = \frac{1}{\numlabeled} \sum_{i=1}^\numlabeled f^2 (x_i)$
be the squared empirical $L^2$-norm, and consider the ball
\begin{align*}
\Ball_\numlabeled(\delta, \Lfun^*) \defn \{f \in \Lfun^* \, \mid \|f\|_\numlabeled <
\delta \}.
\end{align*}
Since the sup norm is stronger than this empirical norm, we have the
sequence of inequalities
\begin{align}
\label{eq:coverupperbound}
\log(N (t; \Ball_\numlabeled(\delta; \Lfun^*); \|\cdot\|_\numlabeled))\leq \log(N (t;
\Lfun^*; \|\cdot\|_\numlabeled)) \leq \log(N (t; \Lfun^*; \|\cdot\|_\Lfun)) =
\order \left(\frac{1}{\epsilon t}\right).
\end{align}

Next consider the $\delta$-localized Gaussian complexity of a function
class $\Fclass$, given by
\begin{equation}
\Gcomp(\delta;\mathcal{F}) = \E_{w}\left[ \sup_{\substack{g\in
      \mathcal{F} \\ \|g\|_\numlabeled\leq \delta}} \frac{1}{\numlabeled} \sum_{i=1}^\numlabeled w_i
  g(x_i)\right],
\label{eq:gaussian_complexity}
\end{equation}
where $\{w_i\}_{i=1}^\numobs$ is an i.i.d.\ sequence of standard
normal random variables. Define the critical radius $\delta_\numlabeled$ as the
smallest $\delta$ that satisfies the \emph{master inequality}
\begin{align}
\label{eq:master_equation}
\Gcomp(\delta; \Lfun^*) & \leq \frac{\delta^2}{2 \sigma}.
\end{align}
With this set-up, it is known~\citep{van2000empirical} that the
$M$-estimator $\fhat_\Lfun$ satisfies a bound of the form
\begin{align}
\label{eq:empiricalerrorbound}
\P \big[ \|\fhat_\Lfun-\fstar\|_n^2 \leq c_1 \delta_\numlabeled^2 \big] \geq 1-
e^{-c_2\frac{\numlabeled\delta_\numlabeled^2}{2\sigma^2}},
\end{align}
where $c_1$ and $c_2$ are universal positive constants.  By Dudley's
entropy integral, the critical radius $\delta_\numlabeled$ is upper bounded by
any $\delta$ which satisfies
\begin{align*}
\frac{1}{\sqrt{\numlabeled}} \int_{\delta^2/2}^{\delta} \sqrt{\log(N (t;
  \Ball_\numlabeled(\delta; \Lfun^*); \|\cdot\|_\numlabeled))} dt \leq
\frac{\delta^2}{\sigma}.
\end{align*}
A little calculation shows that it suffices to choose $\delta_\numlabeled$ such that
\begin{align*}
\delta_\numlabeled^2 & \leq \left(
\frac{\sigma^2}{\epsilon \numlabeled} \right) ^{2/3}.
\end{align*}
Note that this is in fact a global upper bound on $\delta_\numlabeled$, since
the first inequality of~\eqref{eq:coverupperbound} holds for any
setting of the design points $\{x_i \}_{i=1}^\numlabeled$.


\subsection{Proof of the error bound~\eqref{eq:rate_l2}
on $\fhat_{\Hil}$}
As our starting point, we use the master
inequality~\eqref{eq:master_equation} with the shifted function class
$\Lfun^*$ replaced by
$\Hil^* \defn \{f - f^* \, \mid \, f \in \Hil ~,~ \|f\|_{\Hil} \le R\}$.
Recall that $\|f\|_{\Hil}^2 = \int {f'}^2 \pdens^2$ and $R=2$ in our case.
Using known bounds~\citep{mendelson2002geometric} on localized Gaussian complexity
of $\Hil^*$ in terms of the eigenvalues of the kernel $\Ker$, the
master inequality takes the simpler form
\begin{align}
\label{eq:master_eq_bis}
\left( \frac{2}{\numlabeled} \sum_{k,j=0}^\infty
\min\big\{\kereig_{k,j},\delta^2\big\}\right)^{1/2} \le
\frac{R}{\sigma}~\delta^2.
\end{align}
We claim that this master inequality is satisfied by
\begin{align}
\label{EqnSurgery}
\delta_\numlabeled = c_1 \left(\frac{\sigma^2}{R^2 \numlabeled}\right)^{1/3}
\end{align}
where $c_1$ is an absolute constant, independent of $\numlabeled, \epsilon$.
Given this choice, the overall claim follows by applying the non-asymptotic error
bound~\eqref{eq:empiricalerrorbound}.

It remains to show that the choice~\eqref{EqnSurgery} is valid, and we
do so by using the bounds on the eigenvalues from
Lemma~\ref{lem:eigenvalues}.  
For $\delta \in (0,\sqrt{2}/\pi)$, let $k^*,j^*$ be
the largest integers such that $\kereig_{k,j} \ge \delta^2$, or
equivalently such that
\begin{align*}
k/(2\sqrt{2}) + j \epsilon^{-3/4} \le 1/(\pi\delta).
\end{align*}
One can verify that $j^* = \lfloor \frac{\epsilon^{3/4}}{\delta
  \pi}\rfloor$ and $k^* = \lfloor 2\sqrt{2}
\epsilon^{-3/4}\big(\frac{\epsilon^{3/4}}{\delta \pi}- \lfloor
\frac{\epsilon^{3/4}}{\delta \pi}\rfloor\big)\rfloor$. 
We note that for $\delta <\sqrt{2}/\pi <1$, $j^*$ and $k^*$ cannot simultaneously be zero.
Using these cut-off points, the number of eigenvalues $(\kereig_{k,j})$
that are larger than $\delta^2$ is at most $2k_0j^*+k^*+1$.
Moreover, we have
\begin{align}
\sum_{k,j=0}^\infty \min\big\{\kereig_{k,j},\delta^2\big\} \le (2k_0j^*+k^*+1)
\delta^2 + \sum_{k=k^*+1}^{2k_0-1} \kereig_{k,j^*}
+ \sum_{j\ge j^*+1} \sum_{k=0}^{2k_0-1} \kereig_{k,j}.
\label{eq:master_ineq_processed}
\end{align}
Next, we control each term in the above expression. 
For the first term, note that $k_0 \le  \sqrt{2} \epsilon^{-3/4}$ and
$j^* \le \frac{\epsilon^{3/4}}{\delta \pi}$. Moreover, we have  
$k^* \le \frac{2\sqrt{2}}{\pi \delta}$. 
Therefore, the first term is bounded as
$2k_0j^*+k^*+1 \le 4\sqrt{2}/(\delta \pi) +1$. 

As for the second term,
\begin{align*}
\sum_{k=k^*+1}^{2k_0-1} \kereig_{k,j^*} &\le \sum_{k=k^*+1}^{2k_0-1}
\frac{1}{\big(\frac{k}{2\sqrt{2}} + j^*\epsilon^{-3/4} \big)^2\pi^2} \\
&=\frac{1}{\big(\frac{k^*+1}{2\sqrt{2}} + j^*\epsilon^{-3/4} \big)^2\pi^2}
+ \sum_{k=k^*+2}^{2k_0-1}
\frac{1}{\big(\frac{k}{2\sqrt{2}} + j^*\epsilon^{-3/4} \big)^2\pi^2}\\
&\stackrel{(i)}{\le} \frac{1}{\big(\frac{k^*+1}{2\sqrt{2}}+ j^*\epsilon^{-3/4} \big)^2\pi^2}
+ \int_{k^*+1}^{2k_0-1}\frac{dt}{\big(\frac{t}{2\sqrt{2}}+ j^*\epsilon^{-3/4}\big)^2\pi^2}\\
& =  \frac{1}{\big(\frac{k^*+1}{2\sqrt{2}}+ j^*\epsilon^{-3/4} \big)^2\pi^2}
+\frac{2\sqrt{2}}{\big(\frac{k^*+1}{2\sqrt{2}}+ j^*\epsilon^{-3/4}\big)\pi^2}
-\frac{2\sqrt{2}}{\big(\frac{2k_0-1}{2\sqrt{2}}+ j^*\epsilon^{-3/4}\big)\pi^2}\\
&\le \frac{1}{\big(\frac{k^*+1}{2\sqrt{2}}+ j^*\epsilon^{-3/4} \big)^2\pi^2}
+\frac{2\sqrt{2}}{\big(\frac{k^*+1}{2\sqrt{2}}+ j^*\epsilon^{-3/4}\big)\pi^2}\\
& \stackrel{(ii)}{\le} \frac{4\sqrt{2}}{\big(\frac{k^*+1}{2\sqrt{2}}+ j^*\epsilon^{-3/4} \big)\pi^2}.
\end{align*}
The first inequality is obtained using Lemma~\ref{lem:eigenvalues}. 
Inequality $(i)$ follows by upper-bounding the discrete sum by an integral:
$\sum_{k=k^*+1}^{k_0} 1/k^2 \le \int_{k^*}^{k_0} dt/t$ 
(this argument will be used once more to control of the third sum).
Inequality $(ii)$ is obtained simply by factorizing and noting that $\frac{k^*+1}{2\sqrt{2}}+ j^*\epsilon^{-3/4} \ge \frac{1}{2\sqrt{2}}$.
By definition of $j^*$ and $k^*$, we have $ \frac{k^*+1}{2\sqrt{2}}+ j^*\epsilon^{-3/4} > \frac{1}{\pi \delta}$.
Therefore,
\[\sum_{k=k^*+1}^{2k_0-1} \kereig_{k,j^*} \le
\frac{4\sqrt{2}}{\big(\frac{1}{\delta \pi} \big)\pi^2} = \frac{4\sqrt{2}}{\pi}\delta.\]

On the other hand, using the same techniques above, the third term is upper bounded as
\begin{align*}
\sum_{j\ge j^*+1} \sum_{k=0}^{2k_0-1} \kereig_{k,j}
&\le \sum_{j\ge j^*+1} \frac{2k_0}{(j\epsilon^{-3/4})^2\pi^2}\\
& = \frac{2k_0\epsilon^{3/2}}{(j^*+1)^2\pi^2}
+ \sum_{j\ge j^*+2} \frac{2k_0\epsilon^{3/2}}{j^2\pi^2} \\
&\le \frac{2k_0\epsilon^{3/2}}{(j^*+1)^2\pi^2}
+ \int_{j^*+1}^\infty \frac{2k_0\epsilon^{3/2}}{s^2\pi^2} ds \\
&= \frac{2k_0\epsilon^{3/2}}{(j^*+1)^2\pi^2}
+ \frac{2k_0\epsilon^{3/2}}{(j^*+1)\pi^2}\\
&\le 2\frac{2k_0\epsilon^{3/2}}{(j^*+1)\pi^2}.
\end{align*}
Since $j^* = \lfloor \frac{\epsilon^{3/4}}{\delta \pi}\rfloor$,
we have $j^* +1 > \frac{\epsilon^{3/4}}{\delta \pi}$. Therefore,
using $k_0 = \lfloor \sqrt{2} \epsilon^{-3/4}\rfloor$, we have
\[
\sum_{j\ge j^*+1} \sum_{k=0}^{2k_0-1} \kereig_{k,j} \le\frac{4k_0\epsilon^{3/4}}{\pi}\delta \le \frac{4\sqrt{2}}{\pi}\delta.
\]

Putting these estimates together, inequality~\eqref{eq:master_ineq_processed} yields
\[ \sum_{k,j=0}^\infty \min\big\{\kereig_{k,j},\delta^2\big\}  \le (\frac{4\sqrt{2}}{\delta \pi}+1)\delta^2
+ \frac{8\sqrt{2}}{\pi}\delta \le \delta^2
+ \frac{12\sqrt{2}}{\pi}\delta.\]
 
Therefore, the master inequality~\eqref{eq:master_eq_bis} becomes
\[\sqrt{\frac{2}{\numlabeled}} \Big(\delta^2
+ \frac{12\sqrt{2}}{\pi}\delta \Big)^{1/2}
\le \frac{R}{\sigma}\delta^2.\]
Finally, only considering solutions $\delta <1$, it can be verified that the specified choice~\eqref{EqnSurgery} is
adequate, as claimed.

\subsection{Proof of Lemma~\ref{lem:eigenvalues}}

We first characterize the eigenvalues of the kernel $\Ker$ as
solutions to a certain non-linear equation:
\begin{lemma}
The eigenvalues of the kernel $\Ker$ from equation~\eqref{eq:kernel}
are given by the solutions $\lambda > 0$ of the non-linear equation
\begin{align}
\label{EqnNonLinear}
\tan \left(\frac{\epsilon}{ \sqrt{\lambda b}} \right)
\tan\left(\frac{1-\epsilon}{\sqrt{\lambda a}} \right) = \left( \frac{b}{a}
\right)^{3/2}.
\end{align}
\label{lem:double_tan}
\end{lemma}
The proof of this lemma is deferred to the next subsection.  Our next
step is to understand the solutions to
$\tan\big(\frac{\epsilon}{\sqrt{
    b}}x\big)\tan\big(\frac{1-\epsilon}{\sqrt{ a}}x\big) =
\big(\frac{b}{a}\big)^{3/2}$ with $x=1/\sqrt{\lambda}$.

For our purposes, we only need to consider the regime where $b
=\sqrt{\epsilon} \ll a$ and the quantity $\epsilon$ is close to
zero. We then assume that the ratio of the two periods
$\frac{1-\epsilon}{\sqrt{a}}/\frac{\epsilon}{\sqrt{b}} \defn k_0$ is
an integer; note that this assumption can always be satisfied by
choosing $\epsilon$ appropriately. This assumption prevents complicated
oscillatory phenomena, and thus makes the analysis simpler.

Define the function $\phi(x) \defn \tan\big(\frac{\epsilon}{\sqrt{
    b}}x\big)\tan\big(\frac{1-\epsilon}{\sqrt{ a}}x\big)$. We first
exploit the periodicity of the function $\phi$ so as to simplify
the reduce the problem
$\phi$ is even, and by our assumption on $k_0$, it has a period of
$\frac{\sqrt{a}}{1-\epsilon} \pi$. Therefore, we only study its
behavior on $[0,\frac{\sqrt{ b}}{\epsilon} \pi]$.  We divide this
interval into two intervals $I = [0,\frac{\sqrt{ b}}{\epsilon} \pi/2)$
  and $\bar{I} = [\frac{\sqrt{ b}}{\epsilon} \pi/2,\frac{\sqrt{
        b}}{\epsilon} \pi]$ and we study the behavior of $\phi$ on
  each of them separately.

Divide $I$ into smaller subintervals
$I_k \defn \big[k\frac{\sqrt{ a}}{1-\epsilon}
  \pi/2,(k+1)\frac{\sqrt{ a}}{1-\epsilon} \pi/2\big)$ with $0\le k \le
  k_0-1$. On each interval $I_k$, both functions
  $x \rightarrow \tan\big(\frac{1-\epsilon}{\sqrt{ a}}x\big)$
  and
  $x \rightarrow \tan\big(\frac{\epsilon}{\sqrt{ b}}x\big)$
  are continuous, positive, and increasing. In addition, the last one varies
  from 0 to $\infty$. Therefore, $\phi$ spans the entire half line
  $[0,\infty)$ on $I_k$. Consequently, it must cross the line $y =
    \big(\frac{b}{a}\big)^{3/2}$ exactly once in each interval $I_k$.
    We denote the coordinate of intersection by $x_k$, i.e.\
    $\phi(x_k) = \big(\frac{b}{a}\big)^{3/2}$ and $x_k \in I_k$.

Similarly, we divide $\bar{I}$ into regular subintervals
$\bar{I}_k = \frac{\sqrt{ b}}{\epsilon}\pi - I_k$. We observe that by parity of $\phi$,
we have $\phi(-x_k) = \big(\frac{b}{a}\big)^{3/2}$, and by periodicity, $\bar{x}_k = -
    x_k + \frac{\sqrt{ b}}{\epsilon} \pi \in \bar{I_k}$ also verifies $\phi(\bar{x}_k) =
  \big(\frac{b}{a}\big)^{3/2}$.
  The sequence of numbers $x_0 < x_1 <
  \ldots < x_{k_0-1} < \bar{x}_{k_0-1} < \bar{x}_{k_0-2} < \ldots <
  \bar{x}_0$ correspond to the entire set of solutions on the interval
  $[0,\frac{\sqrt{ b}}{\epsilon} \pi]$. Then by periodicity, we obtain
  all positive solutions by translating the above sequence by
  multiples of the period $\frac{\sqrt{ b}}{\epsilon} \pi$.  Therefore
  the eigenvalues of the kernel $\Ker$ form a doubly-indexed sequence
  $(\kereig_{k,j})$ such that
\begin{align*}
\kereig_{k,j} =
\left\{
\begin{array}{ll}
1\Big/\big(x_k+j\frac{\sqrt{ b}}{\epsilon}\pi\big)^2  & k \in [0,k_0-1],\\
1\Big/\big(\frac{\sqrt{ b}}{\epsilon} \pi - x_{2k_0-k-1}+j\frac{\sqrt{ b}}{\epsilon} \pi\big)^2  & k \in [k_0,2k_0-1].
\end{array}
\right.
\end{align*}

Now, we can upper bound $\kereig_{k,j}$ by using the fact $x_k \in I_k$ when either $k$ or $j$ is greater than $1$:
observe that when $k \le k_0-1$, $x_k \ge k\frac{\sqrt{a}}{1-\epsilon}\pi/2$. Likewise, when $k_0 \le k \le 2k_0-1$,
\begin{align*}
\frac{\sqrt{ b}}{\epsilon} \pi - x_{2k_0 - k - 1} \ge \frac{\sqrt{
    b}}{\epsilon} \pi - ((2k_0-k-1)+1)\frac{\sqrt{ a}}{1-\epsilon}
\pi/2 = k \frac{\sqrt{a}}{1-\epsilon}\pi/2.
\end{align*}

Thus, we have shown that $\kereig_{k,j} \le
1\Big/\big(k\frac{\sqrt{a}}{1-\epsilon}/2+j\frac{\sqrt{ b}}{\epsilon}
\big)^2 \pi^2$ for all $1 \le k \le 2k_0-1$.  Furthermore, recalling
that $a = \frac{1/2-\epsilon^{3/2}}{1-\epsilon}$, we notice that
$\frac{\sqrt{a}}{1-\epsilon} \ge 1/\sqrt{2}$ when $\epsilon < 1/2$.
Consequently, we have
\begin{align*}
\kereig_{k,j} \le 1\Big/\Big(\frac{k}{2\sqrt{2}}+j\frac{\sqrt{
    b}}{\epsilon} \Big)^2\pi^2 
\qquad \mbox{
 whenever $k\ge 1$ or $j\ge 1$.}
\end{align*}
Note in passing that $k_0 = \frac{1-\epsilon}{\sqrt{a}}\epsilon^{-3/4}
\le \sqrt{2}\epsilon^{-3/4}$.

The case $k=j=0$ needs extra care. We have $x_0 \in
[0,\frac{\sqrt{a}}{1-\epsilon}\pi/2)$. Since $\phi$ vanishes at 0 and
  is strictly increasing on this interval, it is clear that $x_0
  >0$. Now we proceed by an approximation argument valid in the limit
  $\epsilon \rightarrow 0$, and then invoke monotonicity of the
  solution $x_0$ in $\epsilon$.  For $\epsilon$ sufficiently small and
  by using $b = \sqrt{\epsilon}$, we can uniformly approximate the
  function $\tan\big(\frac{\epsilon}{\sqrt{ b}}x\big)$ by
  $\frac{\epsilon}{\sqrt{ b}}x$ for $0<x< 1$. Then, the equation
  $\phi(x) = \big(\frac{b}{a}\big)^{3/2}$ becomes
  $x\tan\big(\frac{1-\epsilon}{\sqrt{ a}}x\big) = \frac{\sqrt{
      b}}{\epsilon}\big(\frac{b}{a}\big)^{3/2}= \frac{1}{a^{3/2}}$. In
  this regime $a \simeq 1/2$ and the equation can be further
  approximated by $\tan(\sqrt{2}x) = \frac{2\sqrt{2}}{x}$. A numerical
  inspection shows that the latter has a unique solution $.892\le x^*
  \le .899$ on $[0,1]$. Moreover we also numerically observe that the
  solution $x_0(\epsilon)$ to the equation $\phi(x) =
  \big(\frac{b}{a}\big)^{3/2}$ on $[0,1]$ is an increasing function of
  $\epsilon$, hence $x_{0} \ge \lim_{\epsilon \rightarrow 0}
  x_0(\epsilon) = x^*$. Therefore, the first eigenvalue $\kereig_{0,0}
  = 1/x_{0}^2$ is upper bounded by a constant independent of
  $\epsilon$---that is, we have $\kereig_{0,0} \le 1/x^{*2} \le 1.26$,
  as claimed.


\subsection{Proof of Lemma~\ref{lem:double_tan}}

Letting $P$ be the distribution associated with the density $\pdens$,
we study the eigenvalues and eigenfunctions of the integral operator
$T: L_2(P) \rightarrow L_2(P)$ given by
\begin{align*}
Tf(x) & \defn \int_{-1}^1 f(t) \Ker(t,x) \pdens(t)dt.
\end{align*}
Here the reader should recall that the density $\pdens$ takes the form
\begin{align*}
\pdens(x) = \begin{cases} b & \mbox{if $x \in [-\epsilon,\epsilon]$}
  \\
a & \mbox{if $x \in [-1,-\epsilon)\cup(\epsilon,1]$.}
\end{cases}
\end{align*}
where the parameters $(a, b, \epsilon)$ are related by the equations
$(1-\epsilon)a + \epsilon b = 1/2$, and $b = \sqrt{\epsilon}$.  The
kernel $\Ker$ is given by
\begin{align*}
\Ker(x,y) = \frac{1}{4} \int_{-1}^1 \frac{dt}{\pdens^2(t)} -
\frac{1}{2} \left | \int_{x}^y \frac{dt}{\pdens^2(t)} \right |, \quad
\mbox{for $x,y \in [-1,1]$.}
\end{align*}

The eigenvalue equation associated with the operator $T$ can be
written as $T \varphi_\lambda = \lambda \varphi_\lambda$, where
$\varphi_\lambda$ is the eigenfunction associated with the eigenvalue
$\lambda \ge 0$.  Differentiating this equation twice yields a system
of differential equations for the eigenfunctions:
\begin{lemma}
All eigenfunctions of $T$ must satisfy the following system of
differential equations:
\begin{align*}
\lambda b~\varphi_\lambda'' + \varphi_\lambda =0 &~~ \mathrm{on} \quad
        [-\epsilon,\epsilon], \quad \mbox{and} \\
\lambda a~\varphi_\lambda'' + \varphi_\lambda =0 & \quad \mathrm{on}
\quad [-1,-\epsilon )~ \cup ~(\epsilon, 1].
\end{align*}
\label{lem:eig_diff_eq}
\end{lemma}

Solving this system of differential equations yields that any
eigenfunction must be of the form
\begin{align*}
\varphi_\lambda (x) = \begin{cases} A_1 \sin
  \big(\frac{x}{\sqrt{\lambda b}}\big) &~~ \mbox{for $x \in
    [-\epsilon,\epsilon]$,} \\
A_2 \sin \big(\frac{x}{\sqrt{\lambda a}}\big) + B_2 \cos
\big(\frac{x}{\sqrt{\lambda a}}\big) & \mbox{for $x \in
  [-1,-\epsilon)$, and} \\
A_2 \sin \big(\frac{x}{\sqrt{\lambda a}}\big) - B_2 \cos
\big(\frac{x}{\sqrt{\lambda a}}\big) &~~ \mbox{for $x \in (\epsilon,
  1]$,}
\end{cases}
\end{align*}
where we already exploited the fact that $\varphi_\lambda$ has to be
odd. Of course, not all functions of the above form are eigenfunctions
of $T$, since we lost information by taking two derivatives.  In order
to show that $\varphi_\lambda$ is actually an eigenfunction, we need
to verify that it is continuous continuous, and satisfies the
relations $\big(T\varphi_\lambda\big)' = \lambda \varphi_\lambda'$ and
$T\varphi_\lambda = \lambda \varphi$. Actually, the last condition
will be satisfied when the first two are. Together, these conditions
will provide enough constraints to specify the four parameters $A_1,
A_2, B_2$ and $\lambda$ in an unambiguous, modulo a global
multiplicative constant for the first three.

By imposing continuity on the solutions for $\pm \epsilon$, we obtain
an equation relating the parameters of the problem:
\begin{equation}
A_2 \sin \left(\frac{\epsilon}{\sqrt{\lambda a}}\right) - B_2 \cos
\left(\frac{\epsilon}{\sqrt{\lambda a}}\right) = A_1 \sin
\left(\frac{\epsilon}{\sqrt{\lambda b}}\right).
\label{eq:first_eq}
\end{equation}
Next we verify that the condition $\big(T\varphi_\lambda\big)' =
\lambda \varphi_\lambda'$ holds on $ [-\epsilon,\epsilon]$ and
$[-1,-\epsilon) \cup (\epsilon, 1]$ separately.  Since the density
$\pdens$ is even, we have $\Ker(-x,y) =\Ker(x,-y)$ for all $x,y \in
        [-1,1]$. Therefore, for any odd function $f$, we have
\begin{align*}
\int_{-1}^{-\epsilon} f(t) \Ker(t,x) \pdens(t)dt =
-\int_{\epsilon}^{1} f(t) \Ker(t,-x) \pdens(t)dt,
\end{align*}
so one we can study the problem on the interval $(\epsilon, 1]$ and
automatically obtain the corresponding results on $[-1,-\epsilon)$.

\paragraph{Case $x \in (\epsilon,1]$:}
For any odd function $f$, we have
\begin{multline*}
T f(x) = - \frac{1}{2}\int_{-1}^{-\epsilon} a f(t)
\big(\frac{-\epsilon - t}{a^2} + \frac{2\epsilon}{b^2} + \frac{x -
  \epsilon}{a^2} \big) dt -\frac{1}{2} \int_{-\epsilon}^{\epsilon} b
f(t) \big(\frac{\epsilon- t}{b^2} + \frac{x - t}{a^2}\big) dt \\-
\frac{1}{2} \int_{\epsilon}^{1} a f(t) \big(\frac{|t-x|}{a^2}\big)dt.
\end{multline*}
Differentiating with respect to $x$ and setting $f = \varphi_\lambda$
yields
\begin{multline*}
\big(T\varphi_\lambda \big)' (x)= - \frac{1}{2a} \big\{A_2
\big[-\sqrt{\lambda a} \cos \big(\frac{t}{\sqrt{\lambda
      a}}\big)\big]_{-1}^{-\epsilon} + B_2 \big[\sqrt{\lambda a} \sin
  \big(\frac{t}{\sqrt{\lambda a}}\big)\big]_{-1}^{-\epsilon} \big\} \\
 - \frac{1}{2a} \big\{A_2 \big[-\sqrt{\lambda a} \cos
   \big(\frac{t}{\sqrt{\lambda a}}\big)\big]_{\epsilon}^x - B_2
 \big[\sqrt{\lambda a} \sin \big(\frac{t}{\sqrt{\lambda
       a}}\big)\big]_{\epsilon}^{x} \big\} \\
+ \frac{1}{2a} \big\{A_2 \big[-\sqrt{\lambda a} \cos
  \big(\frac{t}{\sqrt{\lambda a}}\big)\big]_{x}^1 - B_2
\big[\sqrt{\lambda a} \sin \big(\frac{t}{\sqrt{\lambda
      a}}\big)\big]_{x}^{1} \big\}.
\end{multline*}
Since we must have $\big(T\varphi_\lambda \big)' = \lambda
\varphi_\lambda$, some algebra then leads to
\begin{equation}
A_2\cos \big(\frac{1}{\sqrt{\lambda a}}\big) + B_2 \sin
\big(\frac{1}{\sqrt{\lambda a}}\big) = 0.
\label{eq:second_eq}
\end{equation}
Equations~\eqref{eq:first_eq} and~\eqref{eq:second_eq} form a linear
system in the coefficients $A_2$ and $B_2$, and solving this system
yields
\begin{align*}
A_2 &= A_1 \frac{\sin \big(\frac{1}{\sqrt{ \lambda a}}\big) \sin
  \big(\frac{\epsilon }{\sqrt{ \lambda b}}\big)}{\sin
  \big(\frac{1}{\sqrt{ \lambda a}}\big) \sin \big(\frac{\epsilon
  }{\sqrt{ \lambda a}}\big)+\cos \big(\frac{1}{\sqrt{ \lambda a}}\big)
  \cos \big(\frac{\epsilon }{\sqrt{ \lambda a}}\big)},\\
B_2 &= -A_1\frac{\cos \big(\frac{1}{\sqrt{ \lambda a}}\big) \sin
  \big(\frac{\epsilon }{\sqrt{ \lambda b}}\big)}{\sin
  \big(\frac{1}{\sqrt{ \lambda a}}\big) \sin \big(\frac{\epsilon
  }{\sqrt{ \lambda a}}\big)+\cos \big(\frac{1}{\sqrt{ \lambda a}}\big)
  \cos \big(\frac{\epsilon }{\sqrt{ \lambda a}}\big)}.
\end{align*}

\paragraph{Case $x \in [-\epsilon,\epsilon]$:} For an odd function $f$,
we have
\begin{multline*}
T f(x) = - \frac{1}{2}\int_{-1}^{-\epsilon} a f(t)
\big(\frac{-\epsilon - t}{a^2} + \frac{x+\epsilon}{b^2}\big) dt -
\frac{1}{2} \int_{-\epsilon}^{\epsilon} b f(t) \big(\frac{|x-
  t|}{b^2}\big) dt \\
- \frac{1}{2} \int_{\epsilon}^{1} a f(t) \big(\frac{\epsilon - x}{b^2}
+ \frac{t-\epsilon}{a^2}\big)dt.
\end{multline*}
Following an argument similar to the previous case, we find that
\begin{multline*}
\big(T\varphi_\lambda \big)' (x)= - \frac{a}{b^2} \big\{ A_2
\big[-\sqrt{\lambda a} \cos \big(\frac{t}{\sqrt{\lambda
      a}}\big)\big]_{-1}^{-\epsilon} + B_2 \big[-\sqrt{\lambda a} \sin
  \big(\frac{t}{\sqrt{\lambda a}}\big)\big]_{-1}^{-\epsilon} \big\} \\
- \frac{1}{2b} \big\{A_1 \big[-\sqrt{\lambda b} \cos
  \big(\frac{t}{\sqrt{\lambda b}}\big)\big]_{-\epsilon}^x + A_1
\big[-\sqrt{\lambda b} \cos \big(\frac{t}{\sqrt{\lambda
      b}}\big)\big]_{x}^{\epsilon} \big\}.
\end{multline*}
Imposing the constraint $\big(T\varphi_\lambda \big)' = \lambda
\varphi_\lambda$ leads to the equation
\begin{equation}
\hspace{-.5cm} A_2 \big(\cos \big(\frac{\epsilon}{\sqrt{\lambda
    a}}\big) - \cos \big(\frac{1}{\sqrt{\lambda a}}\big) \big)+ B_2
\big(\sin \big(\frac{\epsilon}{\sqrt{\lambda a}}\big) - \sin
\big(\frac{1}{\sqrt{\lambda a}}\big) \big) =
\big(\frac{b}{a}\big)^{3/2}A_1 \cos\big(\frac{\epsilon}{\sqrt{\lambda
    b}}\big).
\label{eq:third_eq}
\end{equation}
Finally, plugging the expressions of $A_2$ and $B_2$ in the above
equation and simplifying yields the claimed equation~\eqref{EqnNonLinear}.


\section{Reduction of least squares to constrained optimization}
\label{sxn:least_squares}

In this appendix, we show that in the regime $p \leq d$, minimizing an
objective function that is a weighted sum of a least-squares cost with
a regularization term $\left(\int \|\nabla f(x)\|^p \pdens^2(x)dx
\right)^{1/p}$ will have degenerate solutions, just like the
constrained formulation. Consider a least-squares problem of the form
\begin{align}
\min_f \Big \{ \sum_{i \in \ObsSet} (f(x_i)-y_i)^2 + \lambda R(f) \Big
\},
\label{eq:lsform}
\end{align}
where $R(f)$ is some arbitrary regularization term.  We show that the
above has the same solution as a constrained optimization problem.
Let
\begin{align*}
\fhat = \arg \min_f \Big \{ \sum_{i \in \ObsSet} (f(x_i)-y_i)^2 +
\lambda R(f) \Big \}.
\end{align*}
Then $\fhat$ is equal to the minimizer of
\begin{equation}
\label{eq:cstform}
\min_f R(f) \quad \text{subject to} ~~ f(x_i) = \fhat(x_i), i \in
\ObsSet.
\end{equation}
To see this, suppose that the optimizer of \eqref{eq:cstform} is $g
\neq \fhat$, then it must be that $R(g) < R(\fhat)$ and $\sum_{i\in
  \ObsSet} (g(x_i)-y_i)^2 = \sum_{i\in \ObsSet} (\fhat(x_i)-y_i)^2$,
in which case the function $g$ achieves a smaller value of the
cost~\eqref{eq:lsform} than $\fhat$.

By setting $R(f) = \left( \int \|\nabla f(x)\|^p \pdens^2(x)dx
\right)^{1/p}$, we know from Section~\ref{sec:degenq} that the
solution to the optimization problem~\eqref{eq:cstform} must be
degenerate, and so must the solution to the problem~\eqref{eq:lsform}.


\end{document}